%% file: tpami_main.tex
\documentclass[10pt,journal,compsoc]{IEEEtran}

\usepackage[dvipsnames]{xcolor}
\usepackage{algorithmicx}
\usepackage{algorithm}
\usepackage{algpseudocode}
\usepackage{amsthm,amsmath,amssymb} 
\usepackage{array}
\usepackage{arydshln}
\usepackage{bbding}
\usepackage{booktabs}
\usepackage{caption}
\usepackage{color}
\usepackage{csquotes}
\usepackage{dsfont}
\usepackage{hhline}
\usepackage{hyperref}
\usepackage{graphicx}
\usepackage{mathrsfs}
\usepackage{multirow}
\usepackage{pifont}
\usepackage{scalerel}
\usepackage{slashed}
\usepackage{tabularray}
\usepackage{tabularx}
\usepackage{todonotes}
\usepackage{xspace}

\usepackage[capitalize]{cleveref}

\ifCLASSOPTIONcompsoc
  \usepackage[nocompress]{cite}
\else
  \usepackage{cite}
\fi

\input src/settings
\input src/pre

\begin{document}

\input src/title_and_authors

\input src/headers
\input src/pubid

\IEEEtitleabstractindextext{%
\input src/abstract
\input src/keywords
}

\maketitle

\IEEEdisplaynontitleabstractindextext
\IEEEpeerreviewmaketitle

\input src/main/main

\ifCLASSOPTIONcaptionsoff
  \newpage
\fi

\bibliographystyle{IEEEtran}
\bibliography{references}
\input src/biographies

\end{document}

%% file: src/settings.tex
\newtoggle{nographics}
\newtoggle{lrgraphics}
\newtoggle{nocomments}
\newtoggle{supplementary}

\togglefalse{nographics}
\togglefalse{lrgraphics}
\togglefalse{nocomments}
\togglefalse{supplementary}

\crefname{section}{Section}{Sections}
\crefname{table}{Table}{Tables}
\crefname{figure}{Figure}{Figures}


%% file: src/pre.tex
\newcommand{\Kill}[1]{}
\iftoggle{nocomments}{
    \newcommand{\JH}[1]{\ignorespaces}
    \newcommand{\MN}[1]{\ignorespaces}
    \newcommand{\LA}[1]{\ignorespaces}
    \newcommand{\todo}[1]{\ignorespaces}
    \newcommand{\plan}[1]{\ignorespaces}
}{
    \newcommand{\JH}[1]{\textbf{\textcolor{violet}{SK: #1}}}
    \newcommand{\MN}[1]{\textbf{\textcolor{purple}{DZ: #1}}}
    \newcommand{\LA}[1]{\textbf{\textcolor{orange}{AA: #1}}}
    \newcommand{\plan}[1]{\textsl{\textcolor{gray}{Say: #1}}}
}

\DeclareGraphicsExtensions{.eps,.pdf,.jpg,.png}
\iftoggle{lrgraphics}{
    \graphicspath{{src/media/lr/}{src/media/vector/}{src/media/tables/}}
}{
    \graphicspath{{src/media/hr/}{src/media/vector/}{src/media/tables/}}
}
\makeatletter
\iftoggle{nographics}{
    \LetLtxMacro{\includegraphics@orig}{\includegraphics}
    \RenewDocumentCommand{\includegraphics}{ s O{} m }{%
        {\setlength{\fboxsep}{0pt}%
         \colorbox{lightgray}{\phantom{\IfBooleanTF{#1}{\includegraphics@orig*}{\includegraphics@orig}[#2]{#3}}}%
        }%
    }
}{}
\makeatother

\crefname{section}{Sec.}{Secs.}
\Crefname{section}{Section}{Sections}
\Crefname{table}{Table}{Tables}
\crefname{table}{Tab.}{Tabs.}

\makeatletter 
\DeclareRobustCommand\onedot{\futurelet\@let@token\@onedot}
\def\@onedot{\ifx\@let@token.\else.\null\fi\xspace}

\def\eg{\emph{e.g}\onedot} 
\def\ie{\emph{i.e}\onedot} 
\def\cf{\emph{c.f}\onedot} 
\def\etc{\emph{etc}\onedot} 
\def\wrt{w.r.t\onedot} 

\makeatother

\NewDocumentCommand\iou{}{%
    \ifmmode \text{\textit{mIoU}} 
    \else \textit{mIoU}\xspace 
    \fi %
}
\NewDocumentCommand\semiou{}{%
    \ifmmode \text{\textit{Sem.\,mIoU}} 
    \else \textit{Sem.\,mIoU}\xspace 
    \fi %
}
\NewDocumentCommand\geoiou{}{%
    \ifmmode \text{\textit{Geo.\,mIoU}} 
    \else \textit{Geo.\,mIoU}\xspace 
    \fi %
}
\NewDocumentCommand\georecall{}{%
    \ifmmode \text{\textit{Geo.\,Recall}} 
    \else \textit{Geo.\,Recall}\xspace 
    \fi %
}
\NewDocumentCommand\acc{}{%
    \ifmmode \text{\textit{mAcc}} 
    \else \textit{mAcc}\xspace 
    \fi %
}
\NewDocumentCommand\semacc{}{%
    \ifmmode \text{\textit{Sem.\,mAcc}} 
    \else \textit{Sem.\,mAcc}\xspace 
    \fi %
}



%% file: src/title_and_authors.tex
\title{SSR-2D: Semantic 3D Scene Reconstruction from 2D Images}

\author{Junwen Huang,
        Alexey Artemov,
        Yujin Chen,    
        Shuaifeng Zhi, ~\IEEEmembership{Member,~IEEE,}\\
        Kai Xu,~\IEEEmembership{Senior Member,~IEEE,}
        and~Matthias Nie{\ss}ner,~\IEEEmembership{Member,~IEEE,}
\IEEEcompsocitemizethanks{
\IEEEcompsocthanksitem Junwen Huang, Alexey Artemov, Yujin Chen, and Matthias Nie{\ss}ner are with the Technical University of Munich, Germany. Junwen Huang and Matthias Nie{\ss}ner are also with the Munich Center for Machine Learning~(MCML).
\IEEEcompsocthanksitem Shuaifeng Zhi is with Comprehensive Situational Awareness Laboratory (CSA Lab), College of Electronic Science and Technology, China.
\IEEEcompsocthanksitem Kai Xu is with the Xiangjiang Laboratory,  China.
\IEEEcompsocthanksitem Shuaifeng Zhi (\href{mailto:zhishuaifeng@outlook.com}{zhishuaifeng@outlook.com}) is the corresponding author.
\IEEEcompsocthanksitem Alexey Artemov served as a technical lead for the project.
}
}

%
%

%% file: src/headers.tex
%

\markboth{Submitted To IEEE Transactions on Pattern Analysis and Machine Intelligence}%
{Huang \MakeLowercase{\textit{et al.}}: SSR-2D}

%% file: src/pubid.tex

%% file: src/abstract.tex
\begin{abstract}
Most deep learning approaches to comprehensive semantic modeling of 3D indoor spaces require costly dense annotations in the 3D domain. In this work, we explore a central 3D scene modeling task, namely, semantic scene reconstruction without using any 3D annotations. The key idea of our approach is to design a trainable model that employs both incomplete 3D reconstructions and their corresponding source RGB-D images, fusing cross-domain features into volumetric embeddings to predict complete 3D geometry, color, and semantics with only 2D labeling which can be either manual or machine-generated. Our key technical innovation is to leverage differentiable rendering of color and semantics to bridge 2D observations and unknown 3D space, using the observed RGB images and 2D semantics as supervision, respectively. 
We additionally develop a learning pipeline and corresponding method to enable learning from imperfect predicted 2D labels, which could be additionally acquired by synthesizing in an augmented set of virtual training views complementing the original real captures, enabling more efficient self-supervision loop for semantics. 
As a result, our end-to-end trainable solution jointly addresses geometry completion, colorization, and semantic mapping from limited RGB-D images, without relying on any 3D ground-truth information. 
Our method achieves state-of-the-art performance of semantic scene completion on two large-scale benchmark datasets MatterPort3D and ScanNet, surpasses baselines even with costly 3D annotations in predicting both geometry and semantics.
To our knowledge, our method is also the first 2D-driven method addressing completion and semantic segmentation of real-world 3D scans simultaneously.
\end{abstract}

%% file: src/keywords.tex

%% file: src/main/main.tex
\input src/main/01-intro

\input src/main/02-related

\input src/main/03-method

\input src/main/04-experiments

\input src/main/05-conclusion

\input src/main/06-acknowledgement

%% file: src/main/01-intro.tex
\ifCLASSOPTIONcompsoc
\IEEEraisesectionheading{
\section{Introduction}
\label{sec:introduction}
}
\else
\section{Introduction}
\label{sec:introduction}
\fi
\input src/figures/01-fig-teaser

\IEEEPARstart{D}{igitizing} real-world 3D environments is a multifaceted computer vision problem defined at multiple levels of interpretation. 
It encompasses diverse tasks ranging from geometry and appearance reconstruction (\eg, aiming to realistically reproduce a scene captured on photographs as a textured 3D asset) to predicting semantics and functionality of objects in the scene, to inferring and describing their spatial relationships, to isolation and reconstruction of dynamic changes (\eg,~\cite{slabaugh2001survey,chen20153d,zollhofer2018state,roldao20223d} provide high-level summaries on these and other relevant directions). 
Downstream applications often require jointly exploiting more than a single layer of representation to match their requirements. 
For instance, compelling free-viewpoint virtual tours require objects in the scene to have complete 3D shapes and faithful textures, and change realistically under varying illumination~\cite{chang2017matterport3d,zollhofer2018state}. 
For virtual-real interactions (\eg, such as grasping and manipulation of objects in the virtual world) to be intuitive, individual dynamic entities with their shapes and inherent physical parameters have to be isolated, and more generally, scene layouts should be parsed~\cite{lin2016virtual}.
Similar requirements arise in automatic vision-based robotic grasping~\cite{du2021vision} and robot navigation~\cite{crespo2020semantic}; additional benefits in these tasks could be obtained from leveraging higher level, generalized concepts such as the utility of each entity (\eg, in the form of affordances~\cite{ardon2020affordances}).

However, even for static indoor 3D areas which are the focus of this work, acquisition of high-quality digital replicas remains a challenging endeavor as acknowledged in a variety of recent projects focusing on data collection~\cite{dai2017scannet,straub2019replica,xia2018gibson,chang2017matterport3d}.
Not only are all obtained real-world acquisitions inherently \emph{incomplete} as physical constraints (\eg, occlusions) and limitations of range 3D scanning adversely affect the coverage of 3D areas, but reconstructing additional representations (\eg, 3D semantics) remains an issue\footnote{The ability to more easily produce multiple-representation data (in addition to an increasing photorealism of present computer graphics engines) has most recently pushed multiple groups to curate simulated rather then real-world datasets~\cite{InteriorNet18,xiang2020sapien,roberts2021hypersim,fu20213d,li2021openrooms}.}.
Furthermore, resorting to human experts for either scanning, artistic editing, or building additional layers such as dense annotations is yet unlikely to deliver flawless, complete digital 3D assets while being notoriously labour-intensive~\cite{chang2017matterport3d,dai2017scannet}.

Formed over the recent years, a new significant direction in 3D scanning research aims to produce automatic tools that compensate for scan limitations in a paradigm which can be roughly summarised as \emph{\enquote{observe something and recover the rest}}. 
Within this approach, much effort has been devoted to processing raw inputs of varying types and sparsity, such as single-view~\cite{cai2021semantic,dourado2022data,song2017semantic,wang2019forknet}, multi-view~\cite{dai20183dmv,hu2021bidirectional}, and fused 3D acquisitions~\cite{dai2018scancomplete,hou20193d,dai2021spsg,revealnet}. 
On the other hand, a number of individual target representations have been extensively researched, including geometry completion~\cite{dai2020sg,wang2020deep}, semantic~\cite{dai20183dmv,hu2021bidirectional,kundu2020virtual} and semantic-instance~\cite{hou20193d,revealnet} segmentation, colorization~\cite{dai2021spsg}, \etc. 
All these approaches no longer rely on having complete 3D inputs but instead presuppose that acquisitions are inevitably (and substantially) partial.  
Technically, these works (much like ours) are enabled by developments in 3D deep learning realm which can leverage 3D scan datasets with massive amounts of partial but complementary observations.

From a practical perspective, it is desirable to combine learning multiple meaningful representations in a single algorithm, reasoning beyond purely geometric or textured reconstructions. 
First, solving many tasks in one system would eliminate any additional processing steps that would otherwise have to be taken once a complete reconstruction is obtained, thus making reconstruction approaches easier to apply to large collections of data. 
Second, mutual interplay between complementary tasks could be efficiently leveraged to improve performance across all tasks. 
This has been noted by a number of works targeting semantic scene completion~\cite{song2017semantic,cai2021semantic,dai2018scancomplete,dourado2022data,revealnet} (see \cref{sec:related} for a more complete study of relevant literature) but was not explored beyond reconstructing (color-less) semantic instances.
Conversely, scene reconstruction methods focused on producing increasingly higher quality textured \enquote{3D assets}~\cite{dai2021spsg}, but lacked the capability to generate any additional layers of representation for making sense of the reconstruction.
Lastly, existence of \enquote{a useful structure among tasks} (see, \eg,~\cite{zamir2018taskonomy}) and ways to exploit it has been drawing increasing attention in context of experience transfer and multi-task learning algorithms (ours is an instance of the latter).

In this work, we propose a deep learning-based approach that, for the first time for large-scale 3D scenes, jointly generates a complete, textured 3D reconstruction equipped with an additional layer of useful representation: 3D semantic segmentation. 
Specifically, we jointly predict volumetric geometry, appearance, and semantics using a three-branch 3D neural architecture; to our best knowledge, our system is the first to predict the three complementary targets from imperfect RGB-D scans using a single compact trained model.

Deep convolutional neural networks (CNNs) provide a convenient fit for this purpose, as they can easily fuse and decode multidimensional, heterogeneous signals.
However, training a multi-modal, highly parameterized 3D network in the 3D domain in a densely supervised manner is challenging as it depends on vast amounts of diverse, high-quality, complete, and labeled 3D data. 
Training on synthetic datasets~\cite{song2017semantic,InteriorNet18,xiang2020sapien,roberts2021hypersim,fu20213d,li2021openrooms} can, in principle, provide suitable training data, the resulting models are unlikely to fully generalize to real 3D scans; we decided against this option in our method.
In some instances, real-world RGB-D datasets provide ground-truth 2D/3D semantic segmentation masks (see, \eg, \cite{chang2017matterport3d,dai2017scannet,armeni2017joint}); however, both geometry and semantic labels in these datasets are incomplete and imperfect, contaminating training signal. 
Instead, we opt for a number of design choices enabling our learning algorithm to leverage the original RGB-D images along with 2D semantics only.

First, inspired by recent methods~\cite{dai2020sg,dai2021spsg}, we remove a subset of RGB-D frames from the input and learn to predict a complete semantic scene from an incomplete reconstruction; for this, we design a three-branch deep 3D CNN to jointly output geometry, color, and semantics in each voxel. 
Second, to support end-to-end optimization of our learning-based algorithm, we develop an extended differentiable rendering method, enabling us to render 3D volume data to depth, RGB and semantic images through raycasting directly.
We design our training algorithm to reproduce the original RGB-D data; as a key ingredient for learning semantics, we learn segmentations provided by either manual annotations or a generic neural predictor trained on diverse, multi-domain data. 
Third, we additionally adapt the virtual view augmentation scheme inspired by the recent work~\cite{kundu2020virtual} to further improve training performance given machine-generated imperfect labels.

The direction we chose is in line with recent approaches where 2D view information is used to supervise 3D predictions~\cite{dai2021spsg,genova2021learning}. 
Our method can additionally be viewed as generalizing upon several recent works~\cite{dai2020sg,dai2018scancomplete,dai2021spsg} by integrating 2D RGB and semantic inputs as supervision; in several instances, we compare to these prior arts.

To summarize, our key contributions are as follows:
\begin{itemize}
\item To the best of our knowledge, our approach is the first to address the challenging task of semantic scene completion from incomplete observations of challenging real-world indoor 3D scenes, without requiring manual 3D annotations.

\item We achieve state-of-the-art semantic scene completion performance on two large-scale benchmarks, namely Matterport3D~\cite{chang2017matterport3d} and ScanNet~\cite{dai2017scannet}. 

\item We demonstrate the practicability of our approach in an important special case by supervising it with generic, proxy segmentation labels, without access to expensive human 3D annotations.
\end{itemize}

%% file: src/figures/01-fig-teaser.tex
\begin{figure*}[!t]
\centerline{\includegraphics[width=0.9\textwidth]{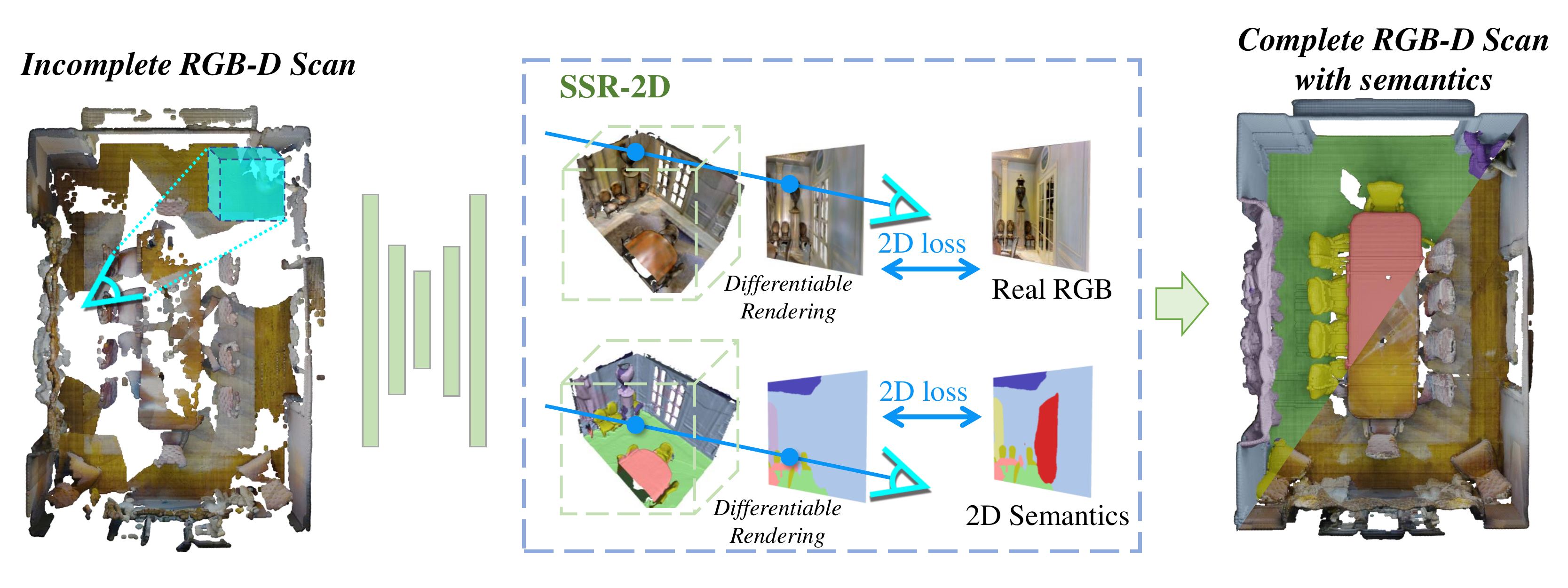}}
\caption{Starting with sparse RGB-D images, our SSR-2D jointly predicts complete geometry, appearance, and semantic labels from incomplete real-world scans, without access to any in-place 3D ground-truth annotations during training.
Instead, we rely on 2D RGB images and their semantic segmentations obtained from a pre-trained semantic predictor.}
\label{fig:teaser}
\end{figure*}

%% file: src/main/02-related.tex
\section{Related Work}
\label{sec:related}

We briefly review closely related approaches that target analysis of large-scale, volumetric 3D scenes, mentioning works for other 3D representations where possible.

\subsection{Semantic Scene Segmentation}
Semantic Scene Segmentation and its variants~\cite{hafiz2020survey,kirillov2019panoptic} generally serve as an initial stage in scene analysis and continue to be extensively researched, in particular for RGB images (see, \eg, \cite{hafiz2020survey,khan2022transformers,zhang2021deep} for a review).
Yet, transferring 2D image segmentation to 3D world is non-straightforward; to this end, specialized point-, voxel-, and mesh-based methods were proposed (surveyed in~\cite{xie2020linking}); we particularly note volumetric approaches~\cite{choy20194d,dai2017scannet,dai2018scancomplete,wang2019voxsegnet} as our network architecturally is a 3D CNN defined on a voxel grid.
Seeking to empower 3D scene segmentation with image-based features, recent approaches propose multiple schemes to leverage 2D and 3D data in parallel~\cite{dai20183dmv,genova2021learning,hou20193d,hu2021bidirectional}. 
Among these, un-projecting per-pixel appearance features from nearby RGB-D images for 3D semantic~\cite{dai20183dmv} and 3D instance~\cite{hou20193d} segmentation, and employing bidirectional view-voxel projection to mutually reinforce 2D and 3D features~\cite{hu2021bidirectional} have proved to be effective. 
For 3D scenes represented as meshes, a more direct approach is to render and segment diversely sampled virtual 2D views using a pre-trained image-based network~\cite{kundu2020virtual}; we adapt their view sampling scheme to our approach.
All these approaches are fully supervised and need dense 3D annotation for training.

\input src/figures/03-fig-method-overview

For reducing labeling costs and aiding generalization for 3D scene understanding, recent works sought to learn representations in a self-supervised fashion in a separate pre-training step (\eg,~\cite{chen2021_4dcontrast,hou2021exploring,xie2020pointcontrast}); however, these methods still require 3D annotations for fine-tuning.
Most similarly to our approach, 2D3DNet~\cite{genova2021learning} obtains 2D features in each image using a pre-trained segmentation model, projects (\enquote{lifts}) these to 3D points, and refines them by a 3D network trained without 3D labels, bypassing the need for 3D annotations during training.
\cite{qian2021recognizing} predicts semantic segmentation for a target viewpoint by rendering a volumetric 3D representation of projected semantics predicted by a pre-trained segmentation model. 
Similarly to the latter two works, one of our experiments uses a pre-trained generic segmentation network. 
All these approaches entirely leave out geometry completion or refinement, instead relying on raw scanned geometry.

\subsection{Scene Completion}
A common requirement in applications is to infer semantic labels not only in directly observed but also in occluded space; to this end, semantic scene completion (SSC)~\cite{song2017semantic} seeks to address both scene occupancy completion and semantic object labeling jointly.
Single-view depth images can be viewed as minimal input data~\cite{guo2018view,song2017semantic,wang2020deep,wang2019forknet,zhang2018efficient,zhang2019cascaded}; alternatives~\cite{cai2021semantic,dai2018scancomplete,dourado2022data,revealnet} (including our method) tackle completing fused reconstructions of entire 3D spaces. 
\cite{dai2018scancomplete} jointly predicts a truncated, unsigned distance field and per-volume semantics in a series of hierarchy levels ranging from low to high resolution.
Leveraging RGB-D image back-projection, \cite{revealnet} combines geometry and appearance features to infer semantics and refine 3D geometry; \cite{dourado2022data} adopts a view of RGB image segmentation as a prior and computes the final semantic scene completion using a 3D CNN.
\cite{cai2021semantic} exploits an interplay between the scene- and instance-level completion tasks and alternates between semantic scene completion and detection of object instances. 
All these methods require dense 3D semantic labels and complete 3D reconstructions during training, making them dependent on synthetic data; in contrast, our algorithm is able to (1) use incomplete, real-world scenes and (2) train from photometric losses obtained via appearance synthesis and segmentations in the 2D domain.

\subsection{Self-Supervised Scene Analysis}
SG-NN~\cite{dai2020sg} learns a scene completion network in a self-supervised learning task where a more complete sparse truncated signed distance field (TSDF) volume is used to guide prediction with inputs from a less complete one. 
Most similarly to our work, SPSG~\cite{dai2021spsg} diverts from using imprecise and incomplete 3D volumetric TSDF values as targets and leverages 2D appearance view synthesis where 2D image-based losses are used for supervising both completion and colorization.
Our work presents an important extension of this line of research by enabling semantic segmentation in addition to generating complete and photometrically realistic scenes.
Recently proposed semantics-enabled variants of neural radiance fields (NeRFs)~\cite{fu2022panoptic,kundu2022panoptic,zhi2021place} strive to represent 3D scene geometry and semantics by a unified neural encoding; unlike these methods, our approach does not require expensive per-scene optimization and can generalize across hundreds of unseen scenes.

%% file: src/figures/03-fig-method-overview.tex
\begin{figure*}[t]
\centering
\includegraphics[width=1\textwidth]{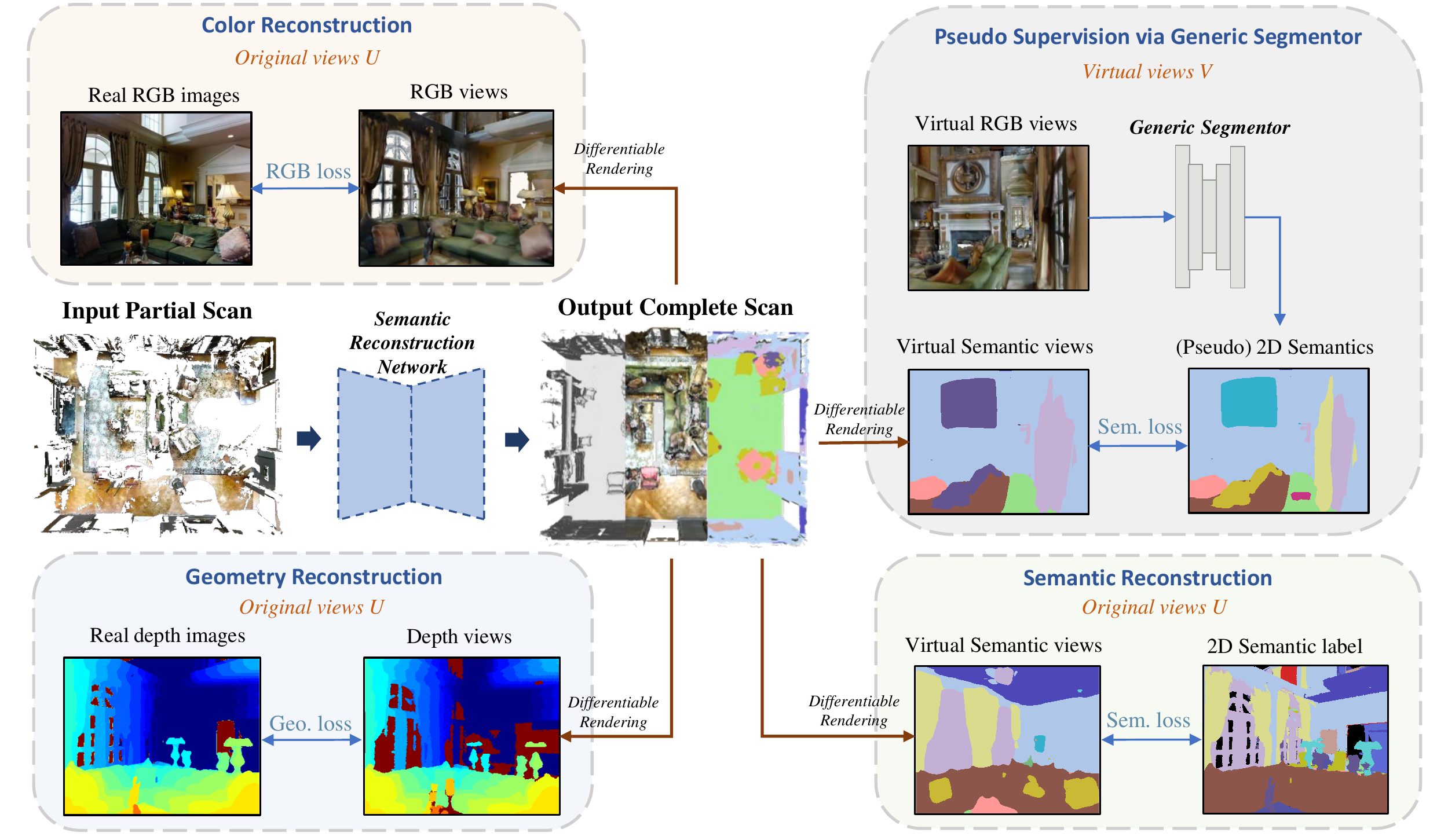}
\caption{Our method accepts a fused but \textit{incomplete} TSDF reconstruction as input and convolves it with a 3D encoder-decoder CNN, jointly producing complete 3D geometry, colorization, and semantic segmentation. 
In the general case (Section~\ref{method:supervision}),
we generate 2D depth, color, and semantic images using either the original viewpoints $U$, or arbitrary virtual viewpoints $V$, by a differentiable rendering technique. 
These synthesized views are used to supervise training \wrt the original RGB-D images in a pseudo-supervised training loop, or \wrt multi-view consistency in a self-supervised training loop.}
\label{fig:method-overview}
\end{figure*}

%% file: src/main/03-method.tex
\section{Method}
\label{sec:method_overview}

\input src/main/03.1-overview

\input src/main/03.2-scene-gen

\input src/main/03.3-diff-rendering

\input src/main/03.4-pseudo_supervision

\input src/main/03.5-virtual-view-generation

\input src/main/03.6-end-to-end-training

%% file: src/main/03.1-overview.tex
\subsection{Method Overview}
\label{method:overview}

The goal of our method is to train a generalizable network~$g$ to perform semantic geometry completion, appearance (color) reconstruction, and semantic labeling simultaneously without having access to any 3D ground-truth (GT) annotations during training.
The input to our method is a set of RGB-D frames $\{(I_u, D_u), u \in U\}$ and their respective estimated camera poses.
To generate input incomplete reconstructions, we select a subset of views $\widetilde{U} \subset U$, fusing these into a truncated signed distance field (TSDF) representation~$d_{\text{in}}$ through volumetric fusion \cite{curless1996volumetric}, of which voxel appearance~$c_{\text{in}}$ is computed by averaging projected pixel color.
As an output, our model predicts a corrected TSDF value~$\widehat{d}$, color~$\widehat{c}$, and semantic label~$\widehat{s}$ in each voxel of the input grid.

Our network~$g$ follows a 3D U-shaped encoder-decoder architecture with two encoders processing geometry and color, and three decoder branches to output geometry, appearance and semantic labels for each voxel, respectively (Section~\ref{method:scene-gen}). 
For computed predictions, we synthesize 2D depth~$\widehat{D}_v$, appearance~$\widehat{I}_v$, and semantic~$\widehat{S}_v$ views via a raytracing-based differentiable rendering process for TSDF volumes~\cite{dai2021spsg} (Section~\ref{method:diff-rendering}).
To enable self-supervised training without ground-truth 3D annotations, we minimize a set of 2D losses involving 
(1) the ground-truth image~$I_v$ and the synthesized color image~$\widehat{I}_v$, 
(2) the reference semantic map~$S^{\text{P}}_v$ (either a ground-truth image, or produced by a generic semantic segmentation model) and the semantic map~$\widehat{S}^{\text{R}}_v$ generated via rendering the reconstructed semantics (Section~\ref{method:supervision}),
(3) the raw depth acquisition~$D_v$ and the depth rendering~$\widehat{D}_v$ of the refined TSDF volume.
For geometry completion, we additionally perform self-supervised training using the incomplete scans to produce their more complete counterparts.

During training, we are able to leverage machine-generated imperfect segmentations from a generic segmentation model~\cite{MSeg_2020_CVPR} instead of GT 2D labels.
We have found that fusing heterogenuous, multi-view information in our learning process results in competitive 3D segmentation performance (Section~\ref{method:pseudo-supervision}) that can be further boosted by integrating generic segmentations computed from a set of virtual viewpoints (\cf, \cite{kundu2020virtual}, Section~\ref{method:virtual-view-gen}).

The overall pipeline of our training procedure is shown in Figure~\ref{fig:method-overview}. 
Our framework is modular, and we have investigated the effect of having various inputs, processing branches, and diverse supervision options; we summarise our conclusions in following sections and report experimental results supporting them in Section~\ref{sec:experiments}.

%% file: src/main/03.2-scene-gen.tex
\subsection{Semantic Scene Reconstruction Architecture}
\label{method:scene-gen}

\input src/figures/03-fig-network-structure

We base our network architecturally on the variant proposed previously for photometric scene generation~\cite{dai2021spsg}.
As input, our algorithm accepts a 4D tensor (\ie, 3D volume with per-voxel TSDF RGB values). 
To extract geometry and color features from the input volumes, we use a dense 3D U-Net~\cite{cciccek20163d} backbone comprised of two 3D encoder branches with 5~ResNet-type~\cite{he2016deep} convolutional blocks each, and three 3D decoder branches with an equal number of convolutional blocks in each.

\noindent \textbf{Network Architecture Details. }
The architecture of our 3D CNN is shown in Figure~\ref{fig:network-structure}, where we summarise the data flow between the layers in our convolutional model.
We extract subvolumes of input data from the scene-level 3D geometry and color data for training purposes. 
First, we process these separately and in parallel by 3D-convolving volumetric inputs with convolutional layers in the two encoders, gradually downsampling the input down to $c_v  \times w \times h \times d$ feature volumes (we use $128 \times 8 \times 4 \times 4$ codes). 
At each block before downsampling, we additionally compute a stacked geometry/color feature map for passing to the respective upsampling blocks in the decoders. 
Next, the independently computed feature volumes are concatenated into a 4D feature map ($2 c_v  \times w \times h \times d$) and fused by a bank of convolutional layers to construct a joint latent feature space with the same shape.
Finally, this fused feature volume is processed by three decoder branches independently to produce refined TSDF~$\widehat{d}$, color~$\widehat{c}$, and semantics~$\widehat{s}$ values in each voxel. 
To enable better feature propagation at each level of the downsampling-upsampling hierarchy in the U-Net, we concatenate the color/geometry encodings to  corresponding upsample layers in the decoders.

\noindent \textbf{Input and Network Varieties without Color Information.} 
We note a few important architectural differences \wrt the original SPSG prototype~\cite{dai2021spsg}, subject to available inputs. 
In certain instances, raw captured inputs may not include color images but provide \emph{depth only} observations~$\{ D_u \}$. 
Importantly, in this regime, colors can potentially assume arbitrary values (\eg, walls in a room may be painted white or vividly textured); however, hallucinating realistic appearance from geometry alone is a difficult generative problem and would require non-trivial complications to our model's architecture (\eg, including an adversarial component in~\cite{dai2021spsg}). 
To verify, we modify our model and make it entirely \enquote{color-blind} by removing 2D and 3D color encoders and decoders, and disabling appearance synthesis during rendering and excluding RGB loss term as well.
Overall, we have found the depth-only inputs to considerably decrease the performance of our method compared to that with RGB-D inputs. We leave detailed specification on training each variant for Section~\ref{method:supervision}.

%% file: src/figures/03-fig-network-structure.tex
\begin{figure}[t]
\centerline{\includegraphics[width=\columnwidth]{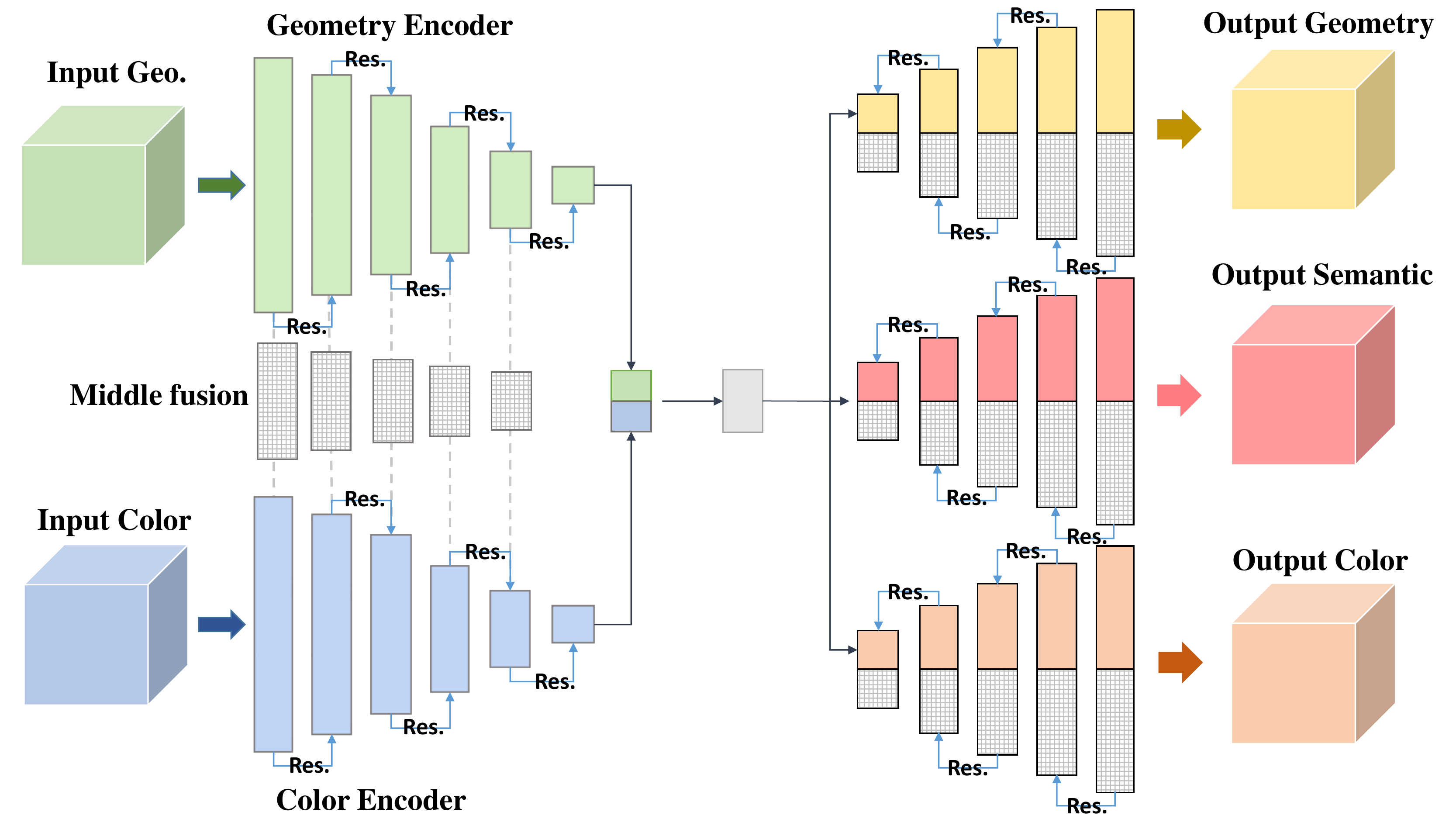}}
\caption{Our 3D CNN architecture comprises two encoders for \colorbox[rgb]{0.82,0.93,0.75}{geometry} and \colorbox[rgb]{0.754,0.830,0.960}{color}, and three decoders for completed \colorbox[rgb]{1.000,0.904,0.603}{geometry}, \colorbox[rgb]{1.000,0.602,0.601}{semantics}, and \colorbox[rgb]{0.973,0.797,0.680}{color}, respectively. 
}
\label{fig:network-structure}
\end{figure}

%% file: src/main/03.3-diff-rendering.tex
\subsection{Differentiable Rendering of Depth, Color and Semantics}
\label{method:diff-rendering}

\input src/figures/03-fig-method-semantic-supervision

Our key design choice is to train a network defined in 3D space, but leverage the information contained in the original 2D RGB-D images (possibly, augmented with semantic information) instead of relying on 3D annotation directly. 
We thus require a 3D-to-2D conversion to enable gradient flow from pixelwise to voxelwise representations. 
Such operations, known as \emph{differentiable volumetric rendering}~\cite{kato2020differentiable}, have proven essential in multiple tasks (\eg, scene generation~\cite{dai2021spsg} or surface reconstruction~\cite{yariv2021volume}).
Among these, we opted to extend a straightforward, efficient raycasting-based rendering approach for TSDF volumes~\cite{dai2021spsg} with a subroutine to render semantic maps.

Our differentiable rendering algorithm~$\mathcal{R}$ accepts a predicted TSDF volume~$\widehat{d}$ with per-voxel predicted colors~$\widehat{c}$ and semantics~$\widehat{s}$ and produces a set $\{\widehat{D}_v = \mathcal{R}(\widehat{d}; v),  \widehat{I}_v = \mathcal{R}(\widehat{c}; v), \widehat{S}_v = \mathcal{R}(\widehat{s}; v) \}$ of depth, color and semantic images, respectively.
To this end, we select RGB-D images taken from the viewing directions $\{v\}$ with the most overlap \wrt the chunk surface (top 5 views each with at least 5\% depth samples within 2\,cm to the near-surface voxels) for supervision.
For rendering semantics, we compute a binary mask to represent each semantic class through raycasting, obtaining a $n_{\text{sem}}$-channel one-hot semantic image where $n_{\text{sem}}$ equals the number of semantic classes. 
Depth and color rendering are obtained using a smilar process to \cite{dai2021spsg}.
The resulting RGB color images contain three channels, and the semantic images contain~$n_{\text{sem}}$ channels, representing binary semantic masks for each of the semantic classes in the class taxonomy of the respective collection~\cite{chang2017matterport3d,dai2017scannet}.
Image resolution of synthesized views is set to $320\times 256$ pixels.

We note that as the predictions are assumed to be complete in 3D, one may use arbitrary viewing directions; we explore this idea in our view synthesis and augmentation technique in Sections~\ref{method:pseudo-supervision}--\ref{method:virtual-view-gen} and its implications for performance in Section~\ref{experiments:setup}.

%% file: src/figures/03-fig-method-semantic-supervision.tex
\begin{figure}[t]
\centerline{\includegraphics[width=.95\columnwidth]{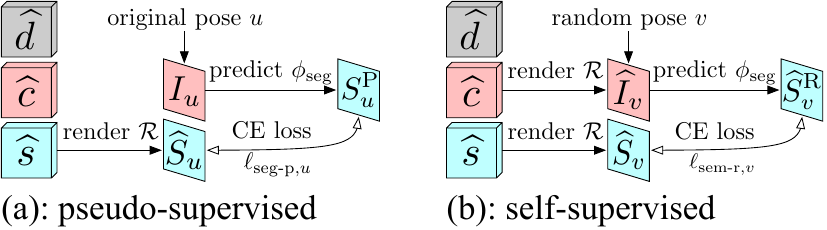}}

\caption{To train our approach with pseudo-GT labels, we optimize: 
(a) for the original views, deviations between segmented RGBs $\phi_{\text{seg}}(I_u)$ and same-view semantic renderings~$\mathcal{R}(\widehat{s}; u)$;
(b) for virtually sampled views, deviations between semantic predictions of RGB renderings~$\phi_{\text{seg}}(\mathcal{R}(\widehat{c}; v))$ and direct semantic renderings~$\mathcal{R}(\widehat{s}; v)$ in the same view.}
\label{fig:semantic_supervision}

\end{figure}

%% file: src/main/03.4-pseudo_supervision.tex
\subsection{Pseudo-Supervision using a Generic Predictor}
\label{method:pseudo-supervision}

To showcase the generalizability and practicality of our method, we discuss a relevant instance of our task~--- complete absence of ground-truth semantic segmentation labels for the original RGB-D data.
Indeed, while obtaining reasonable RGB-D captures is increasingly cheap, their semantic labelling (particularly, manual) remains expensive; the question is then: can we still use our approach to obtain semantic reconstructions in 3D?

We opted to answer this question using a generic semantic predictor (either a pre-trained neural network or an untrained model such as CRFs~\cite{krahenbuhl2011efficient}), which we can use to construct a set of labels, that we refer to as \textit{pseudo-ground-truth}.  
More formally, let $\phi_{\text{seg}}$ denote a function that maps an observed RGB image~$I$ to per-pixel semantic labels~$S$. 
To obtain a set of pseudo-ground-truth semantic labels of captured RGB-D images, we use $\phi_{\text{seg}}$ and compute $S^{\text{P}}_u = \phi_{\text{seg}}(I_u)$. 
As the original RGB images are free from photometric prediction artifacts, using the pseudo-GT $S^{\text{P}}_u$ instead of GT segmentations on source RGB images can provide a stable learning signal. 
In principle, one can view using~$\phi_{\text{seg}}$ as a generalization of annotations for the case of imprecise, universal labels, formulating our pseudo-supervised training loop as shown in Figure~\ref{fig:semantic_supervision}\,(a).

However, since the finite set of source views potentially limits the volume of supervision available to our model, the generic predictor gives us the flexibility to generate pseudo-labels on larger amounts of (synthesized) RGB images from arbitrary poses. 
We thus create a self-supervised training loop, as shown in Figure~\ref{fig:semantic_supervision}\,(b), using only supervision predicted from these virtual views.

To this end, we randomly sample a set of viewing directions~$V = \{v\}$ (see Section~\ref{method:virtual-view-gen} for details of generating these viewpoints), render RGB~$\widehat{I}_v$ and semantic~$\widehat{S}_v$ images for each view, predict a segmentation~$\widehat{S}^{\text{R}}_v = \phi_{\text{seg}}(\widehat{I}_v)$ for a rendered appearance view~$\widehat{I}_v$, and compute per-pixel cross-entropy between pairs~$(\widehat{S}_v, \widehat{S}^{\text{R}}_v)$ of respective semantic maps.
Unlike real captured images, semantic maps $\widehat{S}^{\text{R}}_v$ above need to be produced \textit{on the fly} for rendered RGB images at arbitrary view points. As a result, during training under this setup, we include an extra supervision from this training loop (see~Figure~\ref{fig:semantic_supervision}) with an additive term semantic segmentation cost. 
Integrating both semantic objectives above, we seek to make the directly rendered~$\widehat{S}_v$, pseudo-GT~$S^{\text{P}}_v$, and predicted~$\widehat{S}^{\text{R}}_v$ semantic views differ as little as possible.

%% file: src/main/03.5-virtual-view-generation.tex
\subsection{Virtual View Generation and Selection}
\label{method:virtual-view-gen}

\input src/figures/03-fig-method-view-selection

\input src/tables/table_1.tex

\input src/tables/table_2.tex

We found that generating additional views significantly boosts performance of our method. 
A similar observation has been made by~\cite{kundu2020virtual} who have argued that generating novel views with a wide range of unusual viewing directions and fields of view (FOV) results in a significant improvement in performance of semantic 3D mesh segmentation.
We follow this line of intuition and construct a view selection scheme that we experimentally demonstrate to boost performance of our approach.

We seek to construct virtual views that cover meaningful regions in the scans, observe large number of complete objects, and reflect their contextual relationships.
To this end, we start with the original camera poses which enables binding cameras to room spaces without recomputing chunk-view correspondences, and randomly perturb to their positions, orientations, and FOVs. 
More specifically, we vary \textit{camera intrinsic parameters} (under the pinhole camera model) by enlarging its field of view by a factor uniformly distributed in $[1, 3]$, allowing to capture larger spatial contexts. 
To perturb \textit{camera extrinsic parameters,} we 
\begin{itemize}
\item Randomly disturb the yaw angle of cameras by adding random delta uniformly distributed in~$[-45^{\circ}, +45^{\circ}]$;

\item Randomly disturb the pitch angle of cameras by adding random delta uniformly distributed in~$[-30^{\circ}, +30^{\circ}]$;

\item Randomly offset camera positions by independently shifting along $x, y, z$ axes with a distance distributed uniformly in $[-1\,\text{m}, +1\,\text{m}]$;

\item To enrich the scales of 2D views as well as context information, we randomly translate the camera for up to 2\,m in the direction away from the zero-isosurface of SDFs;

\item Original views are also adopted as they tend to be manually chosen good views within real-world 3d scenes with physical constraints.
\end{itemize}
In practice, we apply combinations of these virtual view generation procedures.
\Cref{fig:method-view-selection} presents example virtual views resulting from our selection process for scenes from Matterport3D~\cite{chang2017matterport3d}.

%% file: src/figures/03-fig-method-view-selection.tex
\begin{figure}[t!]
\centerline{\includegraphics[width=\columnwidth]{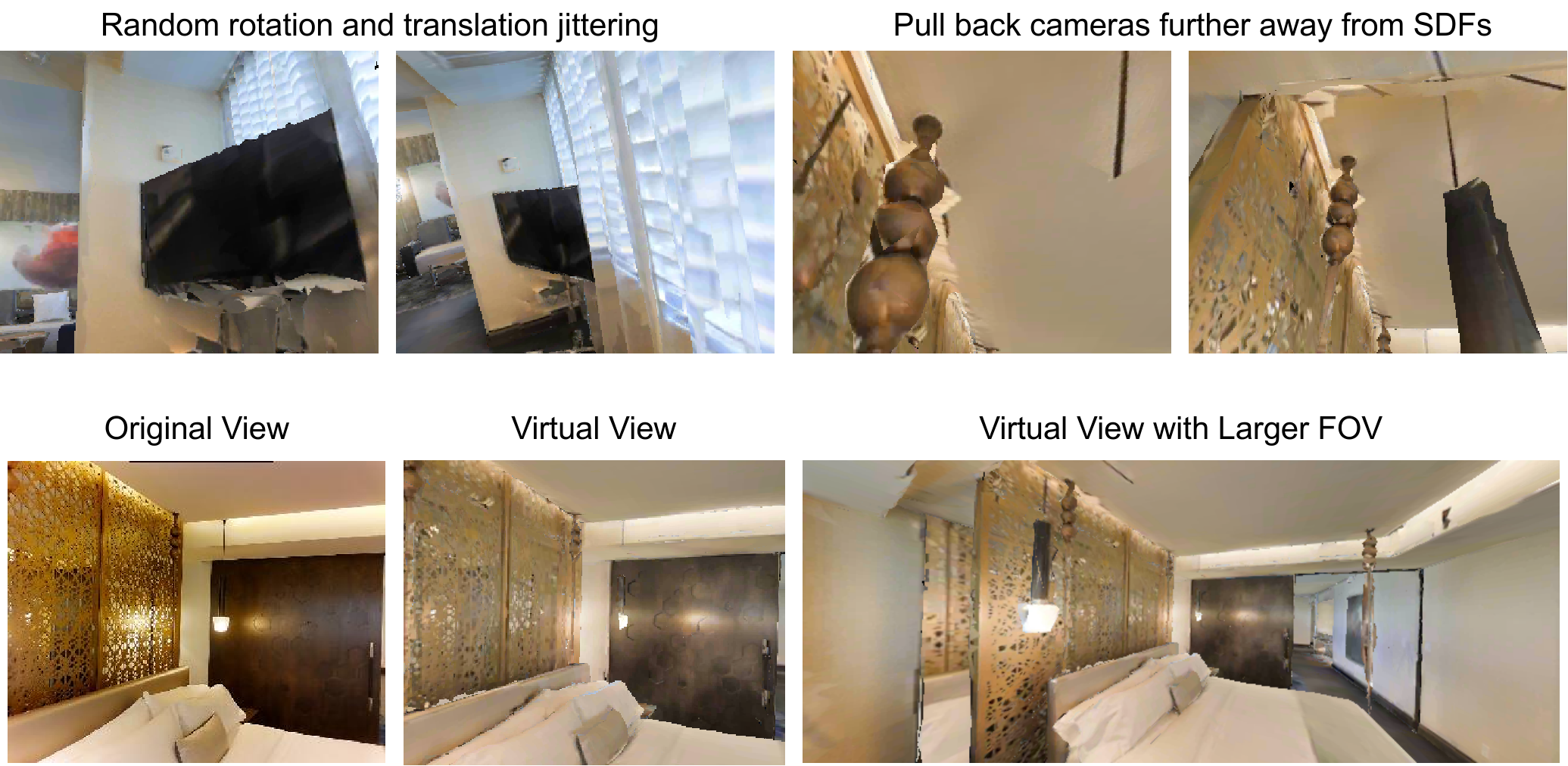}}
\caption{Example virtual views synthesized for the Matterport3D dataset. 
Virtual 2D view selection enables incorporating richer context information of underlying 3D scenes into the renderings for training.}
\label{fig:method-view-selection}
\end{figure}

%% file: src/tables/table_1.tex
\begin{table*}[t]
\centering
\resizebox{0.75\textwidth}{!}{
\begin{tabular}{l ccc cc cc}
\toprule
\multirow{2}{*}{Method} & 
\multicolumn{3}{c}{Supervision} & 
\multicolumn{2}{c}{Matterport3D~\cite{chang2017matterport3d}} & 
\multicolumn{2}{c}{ScanNet~\cite{dai2017scannet}} \\ \cmidrule{2-8}
 & 
\textit{3D GT} & 
\textit{2D GT} & 
\textit{2D Pseudo} & 
\acc & 
\iou & 
\acc & 
\iou \\ \midrule
BPNet~\cite{hu2021bidirectional} &
\Checkmark & 
\Checkmark & 
 & 
12.3 & 
10.9 & 
29.5 & 
22.7 \\
ScanComplete~\cite{dai2018scancomplete} &
\Checkmark & 
 & 
 & 
34.9 & 
28.2 & 
26.9 & 
22.7 \\ 
\textbf{Ours} & 
 &
\Checkmark &
 & 
\textbf{47.9} & 
\textbf{30.1} & 
\textbf{52.0} & 
\textbf{33.3} \\
\hdashline
Ours &
\Checkmark &
 &
 & 
50.0 & 
34.4 & 
64.7 & 
47.1 \\
Ours & 
\Checkmark & 
\Checkmark & 
 & 
53.6       & 
35.1           & 
69.8     & 
51.2      \\
Ours & 
 &
  &
\Checkmark & 
35.3       & 
21.5       & 
37.8    & 
25.9 \\
\bottomrule
\end{tabular}
}

\caption{Semantic scene completion \textit{(SSC)} results for Matterport3D~\cite{chang2017matterport3d} and~ScanNet~\cite{dai2017scannet} benchmarks. 
Our method outperforms two strong baselines and sets the new state-of-the-art on two challenging, real-world benchmarks.
The three bottom rows present varieties of our method leveraging 2D ground-truth \textit{(2D GT)} and/or 3D ground-truth \textit{(3D GT)} labels as well as machine-generated 2D pseudo-labels \textit{(2D Pseudo)} supervision during training (\Cref{experiments:ablative}).}
\label{tbl:semantic-completion}
\end{table*}

%% file: src/tables/table_2.tex
\begin{table}[ht]
\centering
\resizebox{0.825\columnwidth}{!}{
\begin{tabular}{lcc}
\toprule
Method & 
\georecall & 
\geoiou  \\ \midrule
SGNN~\cite{dai2020sg}            & 
57 & 
28 \\
SPSG~\cite{dai2021spsg}  & 
64 & 
39  \\
ScanComplete~\cite{dai2018scancomplete}    & 
36              & 
30            \\
Ours            & \textbf{67.9}   & \textbf{40.2} \\ 
\bottomrule
\end{tabular}
}
\caption{Geometry completion results on Matterport3D~\cite{chang2017matterport3d}. 
Our self-supervised method outperforms all baselines in terms of both measures.}
\label{tbl:geometry-completion-matterport}
\end{table}

%% file: src/main/03.6-end-to-end-training.tex
\subsection{End-to-End Joint Training with 2D Supervision}
\label{method:supervision}

In this section, we introduce adopted loss terms of each key component during training. 
We note that while our learning algorithm is inspired by SPSG~\cite{dai2021spsg}, it differs from SPSG, most importantly, by injecting semantic supervision and excluding adversarial components, which considerably simplifies our system.
Similarly to SPSG though, our final scheme involves the objectives formulated in the \textit{2D domain only,} is end-to-end differentiable, and results in a 3D CNN model.

\noindent \textbf{Geometry Completion. }
To self-supervise the geometry completion task, we use rendered depth images and penalize their deviations to the captured depth maps via a pixel-wise $L_1$ loss:
\begin{equation}
\ell_{\text{geo},u} = 
   \sum\limits_{p} || D_u(p) - \widehat{D}_u(p) ||_1.
\label{eq:geo_loss}
\end{equation}
We additionally use an $L_1$ 3D loss term $\ell_{\text{geo-3d}}$ to self-supervise geometry reconstruction on the predicted 3D TSDF distances, by comparing against their complete counterparts fused from complete RGB-D sequences:
\begin{equation}
\ell_{\text{geo-3d},i} = \frac {1}{n_i} \sum\limits_{p \in \text{chunk}_i} |g(p) - \widehat{g}(p)|_1,
\label{eq:3d-completion-loss}
\end{equation}
where $n_i$ is the number of valid voxels of $i$-th chunk. 
We sum over all chunks to obtain the final 3D loss $\ell_{\text{geo-3d}}$.

\noindent \textbf{Appearance Reconstruction using Raw RGB Images. }
SPSG~\cite{dai2021spsg} heavily emphasizes the need for adversarial training and optimizes a perceptually-based loss for achieving visually compelling RGB synthesis.
In contrast, we opted to train without an adversarial part with either color or normal maps, hence bypassing the need for training a discriminator; to make the training task easier for the optimizer, we additionally exclude the perceptual loss term.
Overall, we have found these modifications to have limited effect on achieving high-quality completion and semantic segmentation while significantly simplifying our system, accelerating convergence, and reducing the number of trainable parameters.
As a result, to enable faithful color synthesis using our model, we simply minimize per-pixel $L_1$ distances between the synthesized appearance view $\widehat{I}_u$ and the target view~$I_u$:
\begin{equation}
\ell_{\text{app},u} = 
  \sum\limits_{p} || I_u(p) - \widehat{I}_u(p) ||_1.
\label{eq:rgb_loss}
\end{equation}

\noindent \textbf{Semantic Segmentation from Real and Virtual Views.} 
Our semantic loss follows a general intuition requiring that the segmentation~$\widehat{s}$ inferred in the 3D domain generates plausible 2D semantic maps~$\{ \widehat{S}_v \}$ under a certain set of 2D views. 
We consider the segmentation labels~$\{ S^{\text{P}}_{\{u,v\}} \}$ available for both the \textit{original captured} RGB-D images $\{ (I_u, D_u), u \in U \}$ as well as for rendered RGB images at \textit{virtual camera poses} $\{ I_v, v \in V \}$ (\Cref{method:virtual-view-gen}), and compute a cross-entropy (CE) loss $\ell_{\text{seg-p}}$ between each pair of rendered~$\widehat{S}_{\{u,v\}}$ and reference semantic views~$S^{\text{P}}_{\{u,v\}}$ (\ie, ground-truth and machine-generated segmentations, the lower index is either~$u$ or~$v$):
\begin{equation}
\ell_{\text{seg-p},{\{u,v\}}} = 
  \sum\limits_{p} L_{\text{CE}}(\widehat{S}_{\{u,v\}}(p), S^{\text{P}}_{\{u,v\}} (p)).
\label{eq:sem_pseudo_loss}
\end{equation}

\noindent \textbf{Joint Training Configuration.}
To summarize, our final training objective integrates geometry, color, and semantic terms 
\begin{align}
L = &\sum\limits_{{\{u,v\}} \in {\{U,V\}}} 
  \frac{1}{n_{\{u,v\}}} 
  \left[ 
    \ell_{\text{geo},u} + 
    \ell_{\text{app},u} + 
    \ell_{\text{seg-p},{\{u,v\}}}
  \right]  \\
  + &\sum\limits_{\text{chunks}} \ell_{\text{geo-3d}}, \nonumber
\label{eq:final_loss}
\end{align}
where, for each batch, the first term sums over the set of $n_{\{u,v\}}$ valid pixels at view set~$u$ or $v$ (\ie, pixels where surface geometry was predicted in~$\widehat{d}$), and the second term sums over number of chunks.
Specifically, to calculate the loss terms in the 2D domain, we use a 3D volumetric mask of the form $\{ x: \widehat{d}(x) < \varepsilon \}$ (we use $\varepsilon = 3\text{\,cm}$) corresponding to the generated geometry, available upon completing geometry in each voxel of the input volumetric grid.

\input src/figures/04.2-fig-semantic-scene-completion-gallery

\input src/figures/04.2-fig-semantic-segmentation-gallery

\input src/figures/04.2-fig-scene-colorization-gallery

%% file: src/figures/04.2-fig-semantic-scene-completion-gallery.tex
\begin{figure*}[htb]
\centerline{\includegraphics[width=\textwidth]{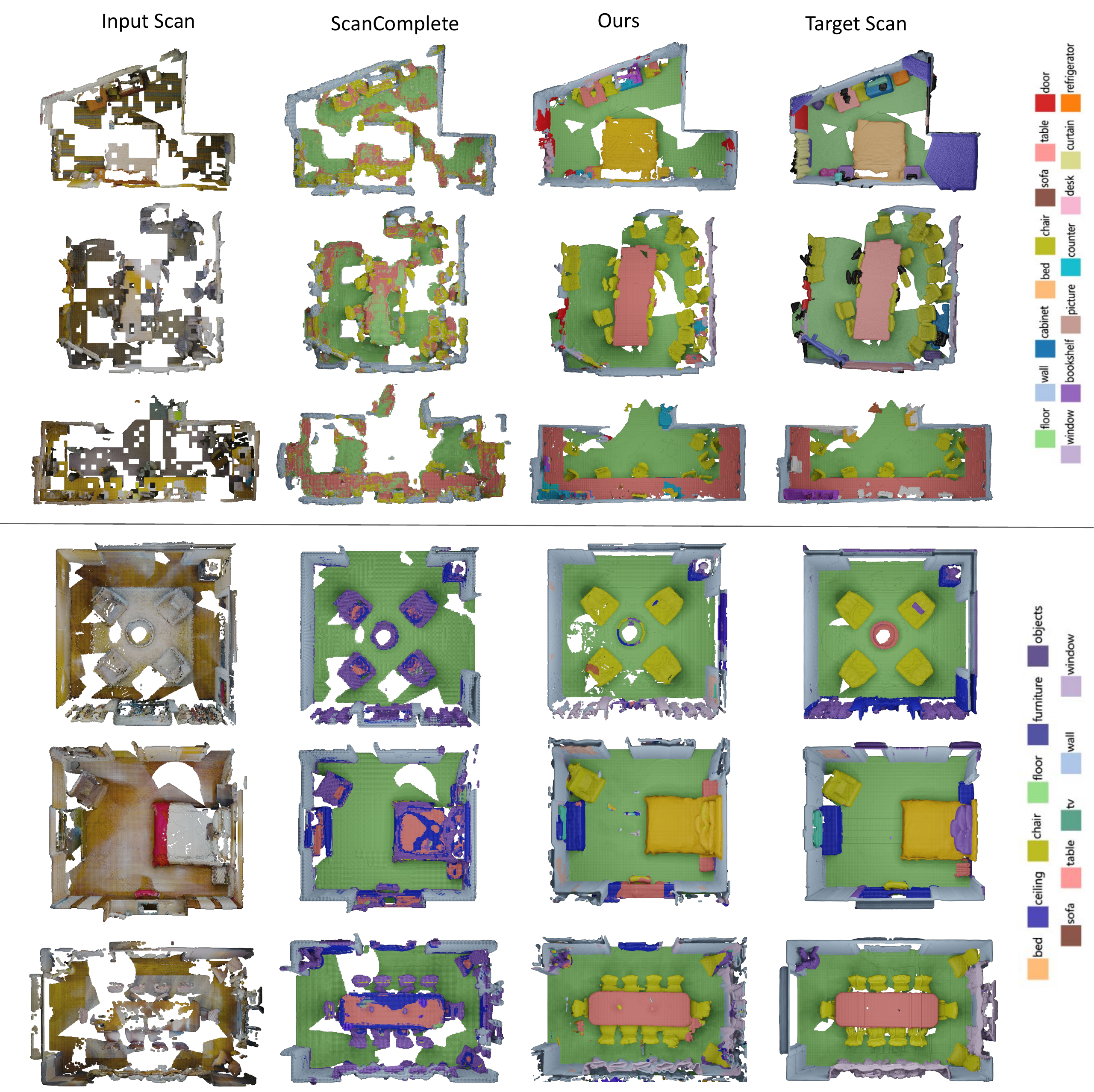}}
\caption{Qualitative semantic scene completion \textit{(SSC)} results using our approach and ScanComplete~\cite{dai2018scancomplete} on ScanNet~\cite{dai2017scannet} (top 3 rows) and Matterport3D~\cite{chang2017matterport3d} (bottom 3 rows) datasets. 
Compared to this baseline, our method predicts more complete geometry and accurate semantics on both input regions as well as unobserved space.}
\label{fig:semantic-scene-completion-gallery}
\end{figure*}

%% file: src/figures/04.2-fig-semantic-segmentation-gallery.tex
\begin{figure*}[t]
\centerline{\includegraphics[width=0.95\textwidth]{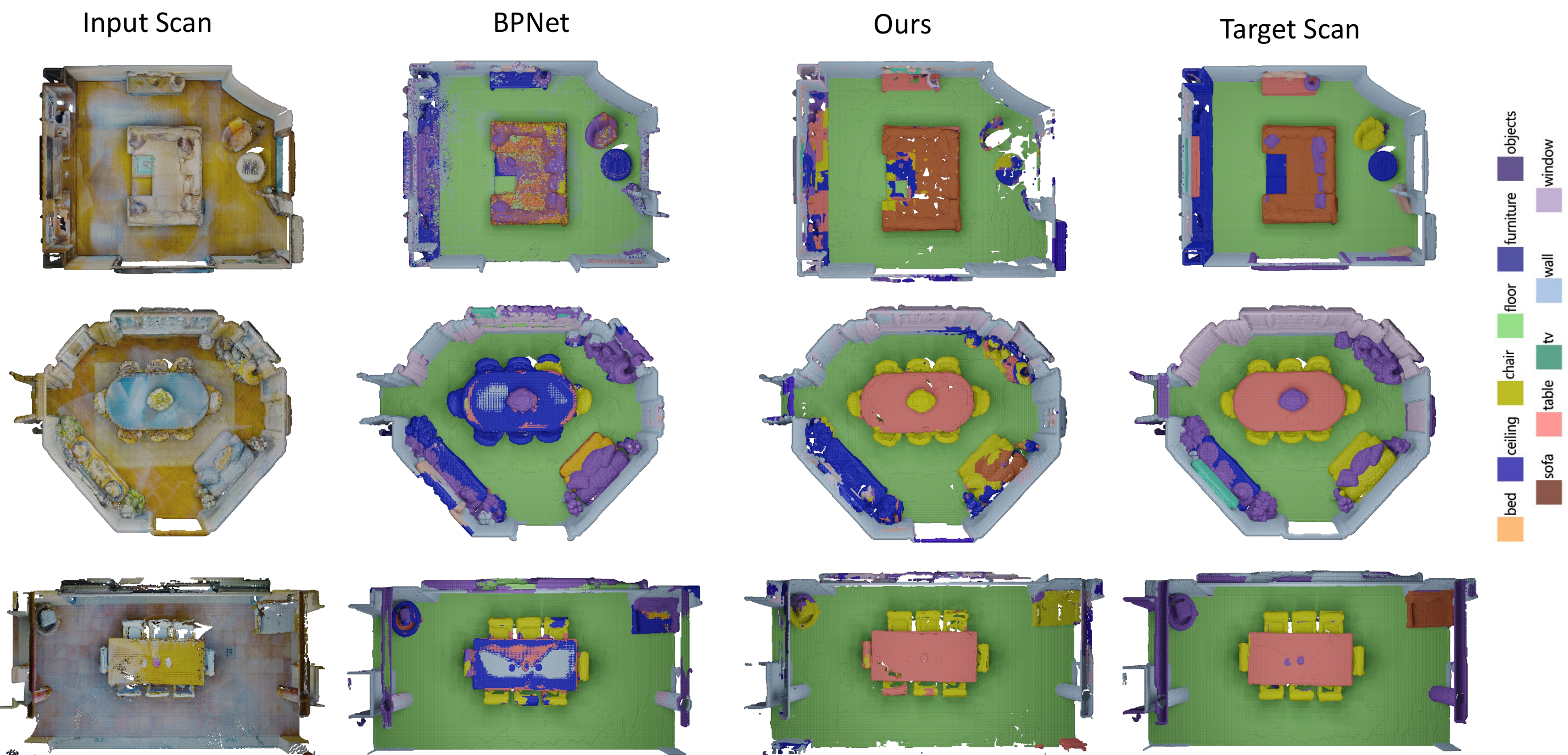}}
\caption{Qualitative semantic segmentation results using our approach and baseline method BPNet \cite{hu2021bidirectional} on Matterport3D dataset. Compared to the baseline, our approach demonstrates robust performance.}
\label{fig:semantic-segmentation-gallery}
\end{figure*}



%% file: src/figures/04.2-fig-scene-colorization-gallery.tex
\begin{figure*}[htb]
\centerline{\includegraphics[width=\textwidth]{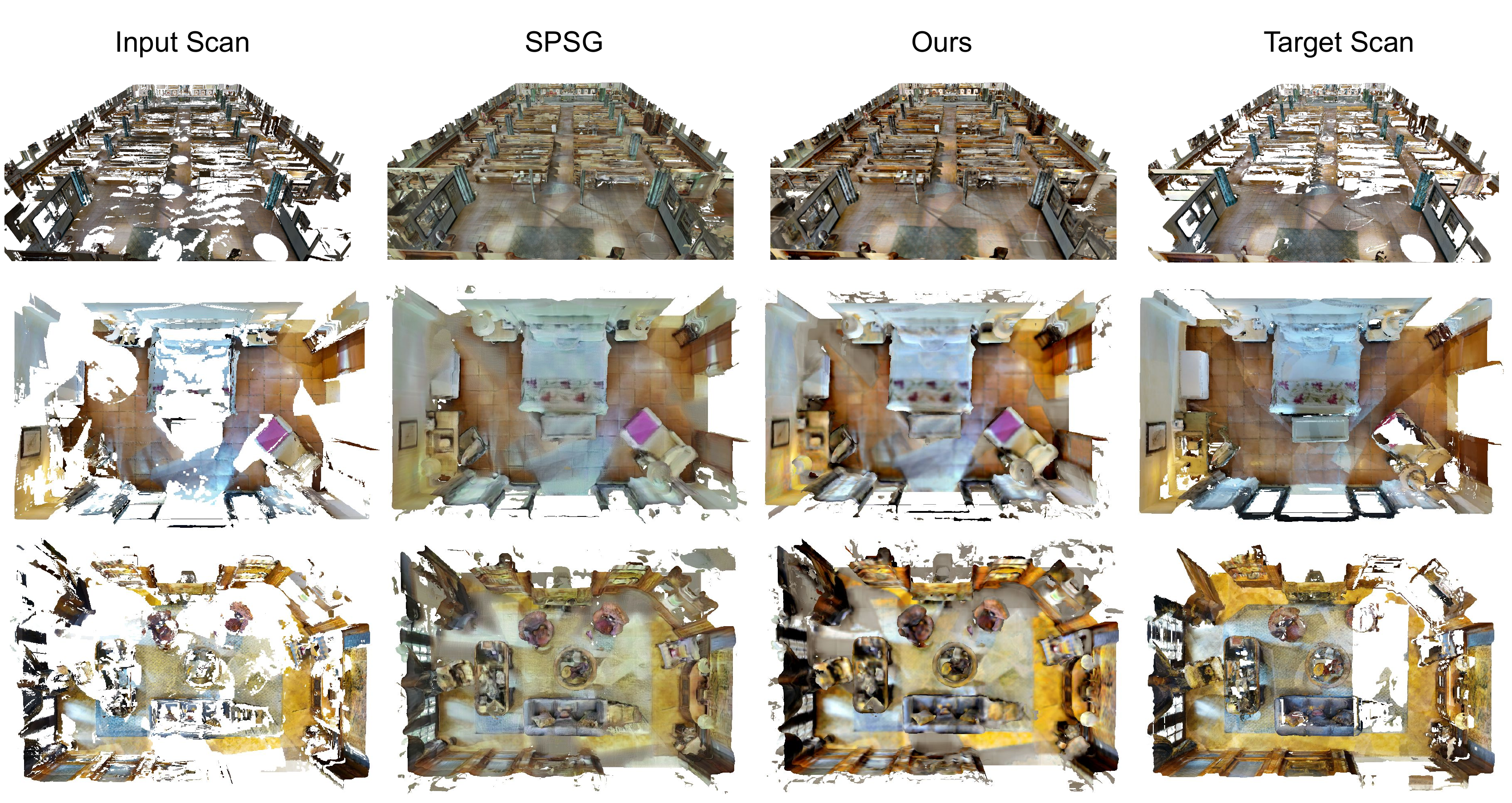}}
\caption{Scene colorization results on the Matterport3D dataset, demonstrating qualitatively comparable performance to SPSG \cite{dai2021spsg} with adversarial components. }
\label{fig:scene-colorization-gallery}
\end{figure*}

%% file: src/main/04-experiments.tex
\section{Experiments}
\label{sec:experiments}

\input src/main/04.1-implementation

\input src/main/04.2-setup

\input src/main/04.3-comparative

\input src/main/04.4-ablative

%% file: src/main/04.1-implementation.tex
\subsection{Implementation Details}
\label{experiments:implementation-details}

\noindent \textbf{Generic Semantic Predictor.}
For training with pseudo-GT labels, we model these using a pre-trained MSeg semantic segmentation network~\cite{MSeg_2020_CVPR}, treating it as a generic semantic predictor ($\phi_{\text{seg}}$ in Section~\ref{method:pseudo-supervision}); we stress that this model is applied in test mode, on data unseen during its training. 
Following~\cite{dai2018scancomplete}, we map the 196 universal classes of MSeg into the most frequent categories in our test data (11~for Matterport3D and 15~for ScanNet). 
Quantitatively, our generic predictor demonstrates a mean IoU performance of 36.9\%  and 37.4\% for Matterport3D and ScanNet, respectively (on the training split without any fine-tuning).

\noindent \textbf{Differentiable Rendering of Semantics. }
Our differentiable renderer produces a multi-channel semantic map in each camera view, in addition to color and depth maps. 
To construct RGB, depth, and semantic channels, we cast rays into each occupied voxel of the scene (a maximum of 640K rays) from the camera's optical center.
We set the maximum depth to 6\,m; pixels corresponding to rays exceeding this depth will be set to a special value.

\noindent \textbf{Training with Virtual View Selection.}
During training, we on-the-fly generate additional, virtual views independent to the original views in our data, to provide auxiliary semantic supervision. 
A similar technique is proposed in~\cite{kundu2020virtual}, yielding improved performance of a semantic 3D mesh segmentation by sampling views with otherwise unusual directions and fields of view.

\noindent \textbf{Training Details.}
We use $128 \times 64 \times 64$ chunks, corresponding to spatial extents of $2.56 \times 1.28 \times 1.28\,\text{m}^3$. 
For training with large-scale scenes in Matterport3D dataset, to enable high-efficiency and data-efficient training, we use a batch size of 2 and the SGD optimizer with an initial learning rate of 0.01, which is annealed according to an exponential learning rate schedule. 
On average, for Matterport3D data, our model needs around 15 epochs (roughly 200K training iterations) and for ScanNet data, 20 epochs (260K iterations) to achieve convergence.
We train our model for about 60\,hours using a single Nvidia RTX3090 GPU.

%% file: src/main/04.2-setup.tex
\subsection{Benchmarks and Data Generation}
\label{experiments:setup}

\noindent \textbf{Benchmark Datasets.} 
To evaluate our system and validate our design choices, we conduct a series of experiments using the challenging large-scale real 3D scans in the Matterport3D~\cite{chang2017matterport3d} and ScanNet dataset~\cite{dai2017scannet} collections.

\textit{ScanNet} is a large-scale real-world RGB-D video dataset providing over 1,500 scenes. Around 2.5 million RGB-D scans are captured by a handheld structure sensor. 
\textit{Matterport3D} provides a vast source of RGB-D scans with 194,400 RGB-D images collected from 90 building-scale scenes. 
Images are collected by a stationary tripod mounted camera rig with three color and three depth cameras, which aims to avoid motion blur and other artifacts sufferred by hand-held video cameras during real-time scanning. 
The different capturing configuration also leads to different processes duing our data generation, as discussed in the next subsection.

Both datasets provide sufficient amounts of real-world training and testing data; following the official guidelines, we use 1788~spaces for training and validation, and 394~for testing on Matterport3D; on ScanNet, we use 1201 spaces for training and 312 for validation.
To obtain highly detailed reconstructions, we use fine voxels with a 2\,cm resolution during TSDF fusion; to enable memory efficient training, we extract $64\times 64\times 128$ subvolumes from fused scans like \cite{dai2021spsg}. 
Overall, we use 77,581 and 88,420 chunks for training with Matterport3D and ScanNet, respectively.

\noindent \textbf{Generation of Incomplete Scans.}
To generate incomplete scans with reasonable instance coverage across scenes for our training and evaluation, we use slightly different procedures for experiments involving Matterport3D and ScanNet collections, given their distinct data acquisition configurations.

For Matterport3D, we generate input reconstructions by selecting a subset of views $\widetilde{U} \subset U$ and fusing these into a TSDF volume using volumetric fusion~\cite{curless1996volumetric}. 
In practice, we use 40\% views on average to build incomplete inputs.

For ScanNet, we use all of the available views to first reconstruct the scene but crop random parts from fused reconstructions by masking the volumetric grid using cube-shaped masks. 
The main motivation behind is that objects within ScanNet scenes are covered by few views, therefore directly fusing a subset of sequences leads to complete loss of certain scene contents which are no longer possible to recover. 
Performing an analysis of the two collections, we have statistically found that, to incompletely sample $\sim$90\% of object instances per scene, ScanNet needs $\sim$60\% randomly chosen frames (in comparison, Matterport3D requires $\sim$38\%), yet keeping this ratio would effectively use nearly all of the data of ScanNet scenes. 
However, aggressively reducing the sampling ratio to match Matterport3D would lead to entirely losing a substantial share of objects in the scene. 
Therefore, we opted for this straightforward, controlled block-masking approach to showcase the capabilities of our method on ScanNet scenes.

As to random masking of ScanNet data generation, we use 3~shapes of cubes, specifically $s_1 = 50^3$, $s_2 = 25^3$, and $s_3 = 10^3$ cubic voxels, respectively (here 1~cubic voxel equals $2\times 2 \times 2$\,cm$^3$). 
We associate each cube with an equal probability $p_i = 0.2, i=1,2,3$, reserving the probability of $1 - \sum_i p_i = 0.4$ for the part of the scan to remain intact. 
To apply masking, we extract subvolumes in the input scan using a sliding window over each of the three dimensions of the volumetric grid by each size of the cube and apply masking by cube shaped as $s_i$ with the probability $p_i$.
As a result, we end up masking nearly 60\% of valid input voxels.

\noindent \textbf{Evaluation Metrics.} 
Following~\cite{dai2020sg,dai2018scancomplete,dai2021spsg,hu2021bidirectional,kundu2020virtual}, for 3D semantic segmentation we report mean intersection-over-union (we use \iou and \semiou interchangeably) and mean voxel-wise accuracy (we use \acc and \semacc interchangeably); 
for geometry completion we report mean geometric intersection-over-union (\geoiou) and mean geometric recall (\georecall).

%% file: src/main/04.3-comparative.tex
\subsection{Comparisons to State-of-the-Art}
\label{experiments:comparative}

\input src/tables/table_3.tex

\noindent \textbf{Evaluation Setups.}
We focus on evaluating two different challenging tasks (1) \textit{semantic scene completion (SSC)} and (2) \textit{semantic scene segmentation (SG)}, as these tasks require predicting different semantic-related quantities. 
In particular, semantic scene completion is our central task as we expect our method to produce high-quality 3D scans with both complete 3D geometry and accurate semantic segmentation, given raw and incomplete RGB-D scans \textit{(SSC)}. 
To separately demonstrate the ability of our method to produce 3D labels given scans of various quality and completeness levels, we additionally evaluate semantic segmentation performance by fixing input geometry and predicting semantic labels in each voxel \textit{(SG)}.

During quantitative evaluation, for SSC task we take the raw, partial scan as input, which in our case has only around 40\% of occupied ground-truth voxels, and evaluate on a complete scene. 
Whereas for a fair comparison of SG task, we exclude the completion part and only evaluate on the voxels that appeared in the input scan. 
To compute accuracy of the SSC task on a certain scene, we divide the number of correct semantic predictions by the number of \emph{ground-truth surface voxels}, where we evaluate the \emph{m-IoU} and the \emph{m-Acc} of all the semantic categories for both input voxels and the ones generated as a completion to the input scan. 


For the SG task, since we focus on semantic segmentation alone which does not change surface geometry, we normalize the number of correct semantic predictions by the number of \emph{input surface voxels}. Importantly, semantic segmentation can be performed in two fine-grained scenarios, that we denote \emph{raw.} and \emph{comp.} (\Cref{tbl:semantic-segmentation}). The former focuses on evaluating semantic segmentation for a significantly incomplete scene, \ie raw, sparse input geometry obtained by fusing a subset of RGB-D images is to be segmented; accordingly, the number of \emph{input surface voxels} of an incomplete scene serves as a normalization constant.
For the latter, \emph{comp.} scenario, we run semantic segmentation on the more complete, ground-truth geometry obtained by fusing all available RGB-D images, normalizing the result \wrt \emph{input surface voxels} viewed as a ground-truth.

As an overall rule to evaluate the output semantics, SSC task takes a partial scan as input but evaluates on a complete ground-truth scan; SG takes either a complete or a partial raw scan as input and evaluates on the same voxels with the input ones. 

\input src/figures/04-fig-nerf-compare

\noindent \textbf{Semantic Scene Completion \textit{(SSC)}.}
As a semantic scene completion baseline, we utilize ScanComplete~\cite{dai2018scancomplete}, a supervised method that operates on 3D TSDF volumes and performs both scene completion and semantic segmentation simultaneously, similar to our proposed method. However, ScanComplete conducts these tasks in a hierarchical fashion. We also evaluate against BPNet~\cite{hu2021bidirectional}, a state-of-the-art supervised method that takes both 3D scans and 2D RGB images as input and supervises the network with semantic labels across domains. Though BPNet does not perform geometric completion, it utilizes 2D domain information to assist 3D semantic prediction. It is noteworthy that these methods leverage 3D annotations as the supervision signal to train a 3D network, which is much more expensive than using only 2D annotations. In contrast, our proposed method can complete the scan geometrically and predict semantics for both input and reconstructed surface voxels, \textit{with only 2D annotations from the given subset of views}, even when given a raw scan with limited RGB-D views. 

In addition, we evaluate the geometric completion results as a reference and compare them to self-supervised SG-NN~\cite{dai2020sg} and SPSG~\cite{dai2021spsg}, which do not aim for the semantic segmentation task. Among these methods, ScanComplete and SPSG perform complex multi-modal training, while SG-NN focuses on completion only.

We present the statistical results in~\Cref{tbl:semantic-completion,tbl:geometry-completion-matterport} on two challenging real-world benchmarks, Matterport3D~\cite{chang2017matterport3d} and ScanNet~\cite{dai2017scannet}. For both the semantic segmentation and geometry completion constituents of the Semantic Scene Completion (SSC) task, our proposed method exhibits a substantial advantage over all baselines on both datasets, demonstrating the effectiveness of our framework which jointly performs appearance prediction, geometry completion, and semantic segmentation in a unified manner. We also provide detailed visual comparisons of semantic scene completion on both datasets in~\Cref{fig:semantic-scene-completion-gallery}.

\input src/tables/table_4.tex

\noindent \textbf{Semantic Scene Segmentation \textit{(SG)}.}
To demonstrate our performance on 3D semantic labeling across various mesh qualities, we also conducted an evaluation of semantic labeling in isolation, without performing geometry completion. Our results for quantitative and qualitative semantic segmentation on the Matterport3D and ScanNet datasets, of which scenes are either partial or complete, are presented in~\Cref{tbl:semantic-segmentation} and~\Cref{fig:semantic-segmentation-gallery}, respectively.

It is important to note that our network was trained jointly for both completion and segmentation, specifically semantic scene completion. To ensure a fair comparison with the state-of-the-art 3D segmentation method BPNet~\cite{hu2021bidirectional}, we utilized the official implementation of BPNet and trained it to predict segmentation labels on input 3D scenes without scene completion. We report the semantic segmentation performance of BPNet by supplying it with the same training input as our method (truncated scans).

Although our method only leverages 2D labels, it outperforms most of the baselines in various aspects. When evaluating complete input scans (\emph{comp.}), our method demonstrates a remarkable improvement compared with all the baselines on ScanNet dataset by a large margin. Similarly, when evaluating the raw partial input scans (\emph{raw}), our method also outperforms all the baselines on the Matterport3D dataset evidently. This indicates that our approach is capable of maintaining high performance on observed 3D regions while generating new, unobserved regions. 

At another end of spectrum, we present the performance given adequate costly 3D annotations to reach the upper bound performance. It turns out that our method could achieve state-of-the-art segmentation performance over all baselines with a single compact model. We believe this result reveals the effectiveness of our framework design with multi-modal learning. We, however, do not claim it as our main result as we are more interested in leveraging more accessible 2D labels.

\input src/figures/04.3-fig-chair

\noindent\textbf{Semantic Fusion via Differentiable Rendering.}
Our method relies on differentiable rendering to enable semantic scene completetion from only 2D images. In this sense, recently proposed Neural Radiance Fields (NeRF) \cite{mildenhall2020nerf} also learns an implicit 3D semantic scene representation from dense 2D observations in a self-supervised manner \cite{zhi2021place}. We think it would be inspiring to see how these methods perform under the challenging setup with sparse RGB-D observations, as both methods aim to semantically reconstruct the 3D scenes from only posed 2D images and labels via differential rendering.
Though our method differs conceptually from NeRF-based approaches such as~\cite{zhi2021place} in a number of ways, making a direct point-to-point comparison less possible and meaningful, 
we conduct quantitative and qualitative comparisons to highlight the generalization capability and crisp reconstruction of complicated indoor scenes by our approach (\Cref{fig:sem-nerf}).

Specifically, we have used the public implementation of Semantic-NeRF \cite{zhi2021place} to evaluate on 5 ScanNet scenes. For a fair comparison, we introduce additional depth supervision into \cite{zhi2021place} and use the same number of input views as ours. 
We empirically found that Semantic-NeRF, with per-scene optimization, struggles to predict sharp 3D geometry for cluttered indoor scenes due to limited views and its density-field representation. To further highlight the performance of semantic label fusion, we isolate the semantic evaluation of \cite{zhi2021place} from its underlying geometry, by projecting its 2D semantic rendering onto the perfect 3D geometry fused from ground-truth depths. As reported in \Cref{tbl:semantic-nerf-compare}, our approach still achieves much better 3D segmentation performance.

\noindent \textbf{Color reconstruction. }
To demonstrate that our model can maintain good quality in both semantic and color reconstructions from input partial scans, we additionally show the qualitative results of the reconstructed colored meshes on Matterport3D dataset in Figure \ref{fig:scene-colorization-gallery}. Compared to SPSG\cite{dai2021spsg}, we achieve equivalent performance on both geometry completion and color inpainting.

%% file: src/tables/table_3.tex
\begin{table*}[htb]
\centering
\resizebox{.85\textwidth}{!}{
\begin{tabular}{lcccccccc}
\toprule
\multirow{3}{*}{Method} & 
\multirow{3}{*}{Input scan} & 
\multicolumn{3}{c}{Supervision} & 
\multicolumn{2}{c}{Matterport3D~\cite{chang2017matterport3d}}     & 
\multicolumn{2}{c}{ScanNet~\cite{dai2017scannet}}  \\ 
\cmidrule{3-9}
   &   
   &
\textit{3D GT} &
\textit{2D GT} &
\textit{2D Pseudo} &
\acc & 
\iou & 
\acc & 
\iou \\ 
\midrule
BPNet~\cite{hu2021bidirectional} & 
\textit{comp.} &
\Checkmark &
\Checkmark &
 &
41.1 &
32.2 &
42.5 & 
33.1 \\
ScanComplete~\cite{dai2018scancomplete}            & 
\textit{comp.}              &
\Checkmark &
 &
 &
42.1 &
29.7 &
44.4 &
30.1 \\
VMFusion~\cite{kundu2020virtual} & 
\textit{comp.} &
& 
& 
\Checkmark &
24.9 &
17.6 &
26.9 &
22.7 \\
Ours & 
\textit{comp.} &
&
\Checkmark &  
&
39.3 &
28.2 &
52.2 &
37.6 \\ 
\hdashline
Ours & 
\textit{comp.} &
\Checkmark &    
&      
&
\textbf{46.3} &
\textbf{33.5} &
\textbf{62.8} &
\textbf{43.4} \\ 
Ours                    & \textit{comp.}              &     &       &\Checkmark               &32.9                   &20.4                      &39.1                     &24.9         \\ 
\hhline{=========}
BPNet~\cite{hu2021bidirectional}                    & 
\textit{raw} &
\Checkmark &
\Checkmark  &  
&
47.7 & 
33.3 &
\textbf{68.2} &
\textbf{57.6} \\
ScanComplete~\cite{dai2018scancomplete}            & \textit{raw}              &\Checkmark       &       &             & 46.6                  & 35.8                    &44.4                     &30.1       \\
\textbf{Ours} & 
\textit{raw}  &   
&
\Checkmark &  
&
48.3 &
36.7 &
50.8 &
36.4 \\ 
\hdashline
Ours                    & \textit{raw}              &\Checkmark       &       &             &\textbf{50.9}                   &\textbf{39.5}                     &64.6                     &48.2         \\
Ours                    & \textit{raw}              &       &       &\Checkmark             &35.3                   &21.5                     &37.8                     &21.9         \\ 
\bottomrule
\end{tabular}
}
\caption{Semantic scene segmentation \textit{(SG)} with partial \textit{(raw)} and complete \textit{(comp.)} input scans for Matterport3D~\cite{chang2017matterport3d} and~ScanNet~\cite{dai2017scannet} benchmarks. 
In both sections, two bottom rows present varieties of our method leveraging 3D ground-truth \textit{(3D GT)} labels as well as machine-generated 2D pseudo-labels \textit{(2D Pseudo)} supervision during training (\Cref{experiments:ablative}).
}
\label{tbl:semantic-segmentation}
\end{table*}

%% file: src/figures/04-fig-nerf-compare.tex
\begin{figure}[t]
\centerline{\includegraphics[width=\columnwidth]{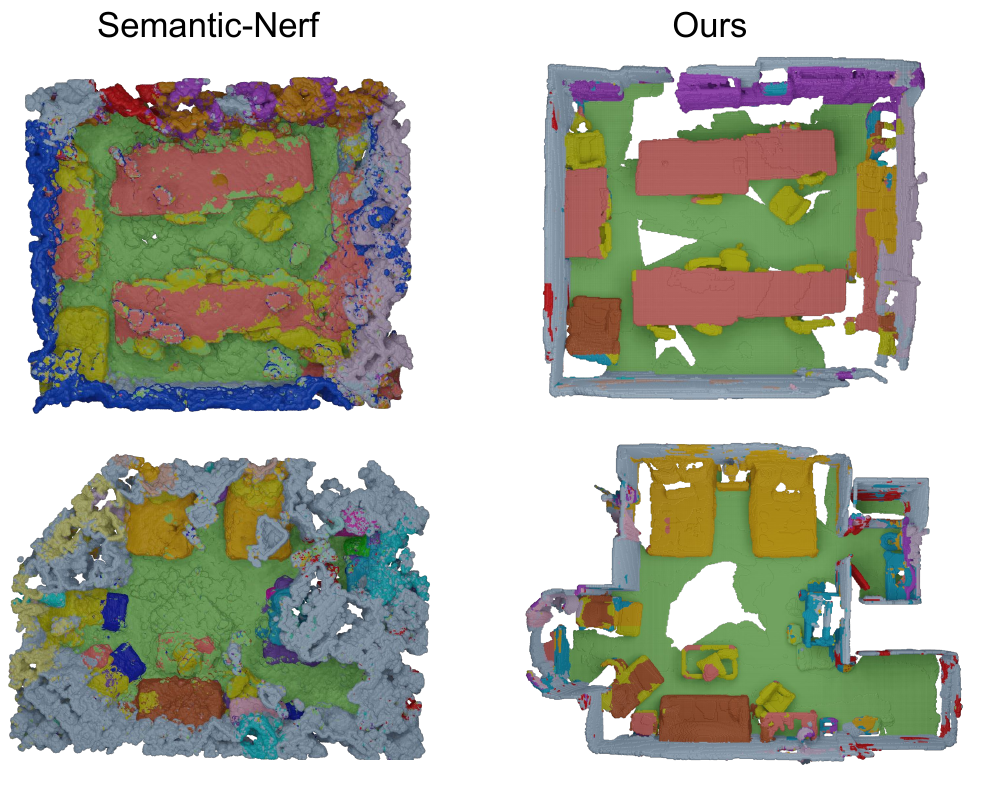}}
\caption{Qualitative semantic reconstruction \textit{(SSC)} results using our approach and baseline method (\ie, depth-aided Semantic-NeRF~\cite{zhi2021place}) on the ScanNet dataset. }
\label{fig:sem-nerf}
\end{figure}

%% file: src/tables/table_4.tex
\begin{table}[t]
\centering
\resizebox{0.99\linewidth}{!}
{
\begin{tabular}{lcccc}
\toprule
\multirow{2}*{Method} & 
\multicolumn{2}{c}{ScanNet~\cite{dai2017scannet}} & 
\multicolumn{2}{c}{Matterport3D~\cite{chang2017matterport3d}} \\ \cmidrule{2-5} 
& 
\semiou & 
\geoiou & 
\semiou & 
\geoiou \\ 
\midrule
Semantic-NeRF~\cite{zhi2021place}        & 54.6               & 3.9               & 49.5               & 2.5               \\
Ours                 & \textbf{62.9}      & \textbf{49.8}     & \textbf{57.9}      & \textbf{46.3}     \\ \bottomrule
\end{tabular}
}
\caption{\textit{SSC} results vs. Semantic-NeRF~\cite{zhi2021place}. Note that unlike this method, our algorithm delivers an accurate, clean geometric reconstruction (see Figure~\ref{fig:sem-nerf}).}
\label{tbl:semantic-nerf-compare}
\end{table}

%% file: src/figures/04.3-fig-chair.tex
\begin{figure}[tb]
\centerline{\includegraphics[width=\columnwidth]{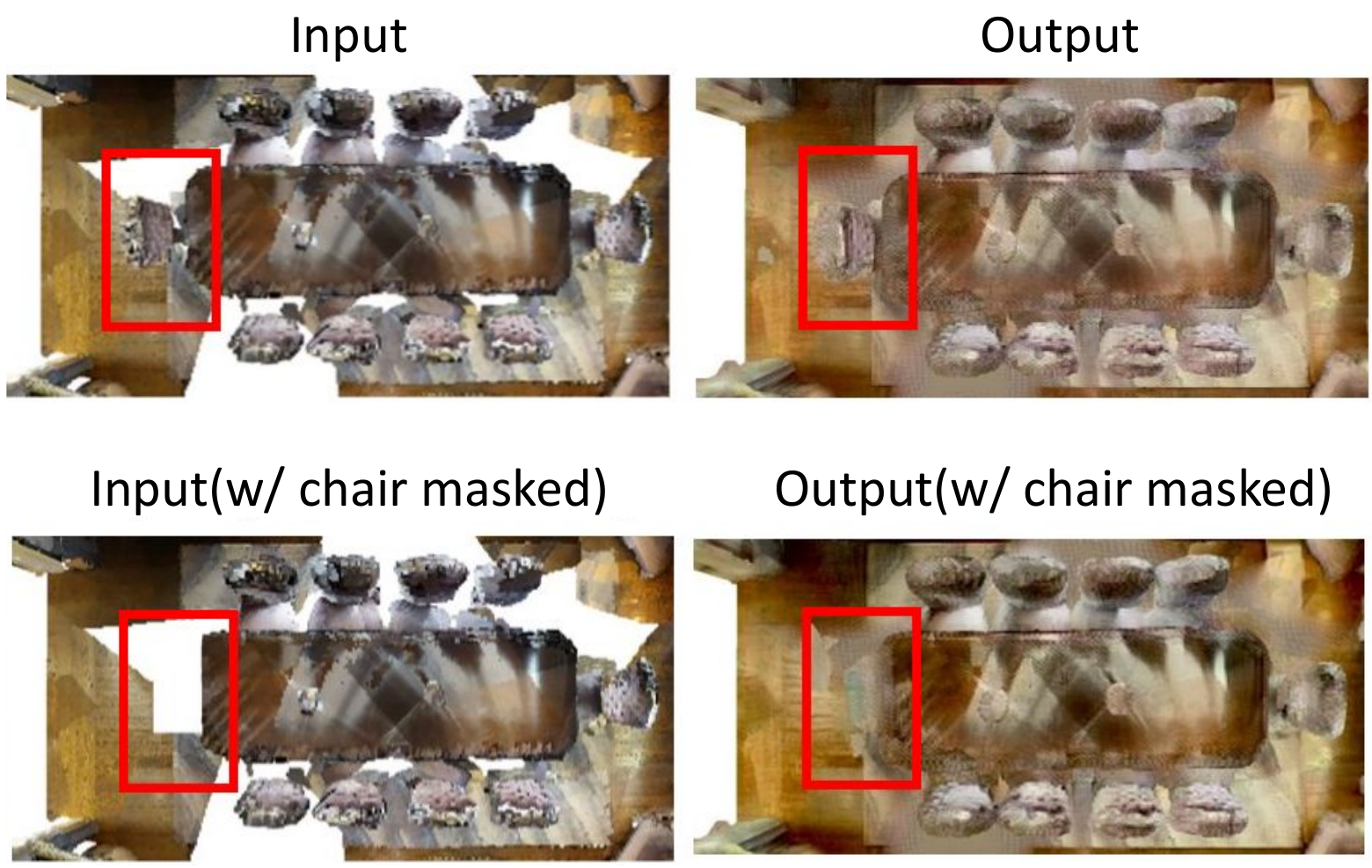}} 
\caption{An ablative visualization for an unobserved chair, showing that our method can complete the scene from the real-world context reasonably, but will not give fabricated predictions if no contextual information is provided.}
\vspace{-0.3cm}
\label{fig:chair}
\end{figure}

%% file: src/main/04.4-ablative.tex
\subsection{Ablative Studies}
\label{experiments:ablative}

\noindent \textbf{Effect of Pseudo Labeling. } 
Within this section, we undertake a rigorous quantitative analysis to evaluate the performance implications of substituting 2D GT labels with machine-generated 2D pseudo-labels, derived from a generic network trained using Equation~\eqref{eq:sem_pseudo_loss}. The results presented in \Cref{tbl:semantic-segmentation} highlight the remarkable superiority of our method over the direct fusion technique VMFusion~\cite{kundu2020virtual}, which also relies exclusively on 2D labels for the reconstruction of 3D semantics. It should be noted that the reported results correspond to the semantic segmentation (SG) task conducted on complete input scans, as VMFusion necessitates an adequate array of views to reconstruct the entire scene.

Given the unavailability of a publicly accessible implementation of VMFusion~\cite{kundu2020virtual}, we present its performance by fusing the same set of training 2D labels into 3D voxels, utilizing the camera poses and depths provided by our method. It is crucial to emphasize that, in order to ensure a fair and unbiased comparison, we meticulously implemented VMFusion~\cite{kundu2020virtual} and employed identical 2D pseudo-semantic labels during the inference process, mirroring our own method. The resulting 3D segmentation is obtained through a process of back-projection and fusion, wherein the 2D pseudo labels are combined with ground truth depths and camera poses. This further underscores the effectiveness of our approach in bridging the gap between 3D and 2D observations, particularly when contrasted with purely view-based methodologies.

Furthermore, we report the results of the Semantic Scene Completion (SSC) task in~\Cref{tbl:semantic-completion}, which exhibits a decline in performance compared to our model trained with 2D GT labels. Nonetheless, the achieved results still remain reasonable when compared to the outcomes of SSC baselines. These findings elegantly demonstrate the inherent flexibility of our framework in effectively leveraging imperfect generic labels, thereby showcasing its potential benefits.

\input src/tables/table_5.tex

\noindent \textbf{Effect of Virtual View Selection.}
We delve into the investigation of the influence of utilizing varying quantities of synthetic views generated by our virtual view generation and selection mechanism (described in Section~\ref{method:virtual-view-gen}). We provide a comprehensive analysis of the number of virtual views supervised using both 2D GT labels and 2D pseudo labels in~\Cref{tbl:ablation-virtual-views}. For 2D GT supervision, we leverage 3D annotations to render virtual views corresponding to specified poses, thus providing additional supervision signals from the 2D domain; for 2D pseudo supervision, we leverage the pre-trained predictor to produce the segmentation labels for the virtual RGB images.

It is important to note that we limit the number of additional views for each chunk to five in order to ensure fair comparisons with the baselines and optimize training efficiency. We observe that a greater number of additional views leads to improved results in terms of semantic reconstruction. However, in the case of pseudo-label training, the performance is constrained by the limited quality of the pre-trained and re-mapped labels, preventing the attainment of exceptionally high accuracy. Nevertheless, we believe that incorporating a larger number of views based on 2D ground truth labels will yield improved performance accordingly, gradually reducing the performance gap compared to GT labels (\Cref{tbl:ablation-virtual-views}).

\input src/tables/table_6.tex

\noindent \textbf{Effect of Geometric Completion on Segmentation.}
Although not the primary focus of our paper, it is pertinent to inquire whether learning scene completion is instrumental when the objective is solely semantic segmentation of potentially incomplete input meshes. To address this question, we conducted an experiment by removing the completion head from our framework. As shown in~\Cref{tbl:ablation-sg-geometry-completion}, there is a significant decline in semantic segmentation accuracy without the inclusion of completion. Hence, we draw the conclusion that explicit and joint reasoning of geometry and semantics is advantageous for semantic segmentation, particularly when dealing with challenging partial 3D structures.

\noindent \textbf{Effect of Color Supervision.}
At the other end of the spectrum, we examine the impact of color information on the attainment of precise semantic reconstruction outcomes. In our investigation, we conduct an ablation study on the network architecture depicted in~\Cref{fig:network-structure} by selectively removing either the color encoder branch or the color decoder branch within the framework of the SSC task. 

Intuitively, the role of the color encoder is to capture RGB information from the raw partial input scan, thereby contributing to the extraction of textural and contextual features for the latent embedding. Conversely, the color decoder functions as a multi-task header in conjunction with the geometric completion and semantic prediction branches. It is expected that the actual RGB images provide supervision for the color reconstruction of the predicted voxels.

According to the findings presented in~\Cref{tbl:ablation-ssc-color}, the inclusion of color information in either the encoder or decoder is vital for the performance of our approach. Moreover, we observed that the absence of the color encoder leads to a more significant decline in performance compared to the absence of the color decoder. This suggests that encoding color information into the feature embedding is of greater importance for the prediction of both semantic completion. Thus, we conclude that, in our case, the multi-modal training benefits more from incorporating color information at the input stage rather than relying solely on supervision from the output head.

\input src/tables/table_7.tex

\noindent\textbf{Completing an Absent Object.} \Cref{fig:chair} demonstrates the capacity of our methodology to effectively complete chair geometry using partial cues, while highlighting the challenges associated with inferring a chair in the absence of any geometric cues. We select a scan crop where a chair is only partially observed (upper right input) and utilize this scan as input to our model, resulting in the generation of a complete chair (upper right output). To facilitate comparative analysis, we subsequently mask the chair in the input scan(lower right input), consequently, the chair's presence is entirely removed, as evident in the output scan (lower right output) where the chair is conspicuously absent. This outcome is consistent with our expectations, as our network is trained to proficiently \textit{complete} unobserved regions within the context of observed contextual information, rather than engaging in holistic scene \textit{generation}. By prioritizing the network's capacity to generate coherent content specifically tailored to real-world scans, we ensure that the network does not engage in the production of arbitrary or unrealistic content.

\noindent \textbf{Effect of 3D-2D Joint Training.}
In order to assess the upper limit of performance achievable by our method, we leverage both 2D and 3D ground-truth annotations as the supervision signal during network training. The outcomes of the semantic scene completion task are presented in \Cref{tbl:semantic-completion}, revealing that the inclusion of supervision from both modalities yields enhanced performance when compared to utilizing solely 2D or 3D ground-truth labels. This finding emphasizes the versatility of our framework in accommodating diverse forms of supervision, thereby demonstrating its scalability across different modalities.

\noindent \textbf{Robustness Concerning Different Levels of the Sparsity of Input Scans.} To assess the resilience of our model, we conducted experiments involving incomplete input reconstructions by selectively using different proportions of views from the captured real images. In Table~\ref{tbl:sparsity-of-inputs} above, we present the impact of varying degrees of input data completeness during the inference process. Our findings reveal that our method consistently performs well when provided with 30\% to 50\% of the frames. Even when reducing the percentage of available input frames to as low as 10\%, the performance degradation remains within reasonable bounds.

\input src/tables/table_8.tex

\noindent \textbf{Cross-Evaluation on Real-World Datasets.} To assess the generalizability of our method beyond trained scenes, we conduct cross-evaluation experiments using two distinct real datasets: Matterport3D and ScanNet. The purpose is to examine how our model scales across different domains with varying room scales and structures. The results, as summarized in Table~\ref{tbl:cross-validation}, demonstrate that our model exhibits the capability to generalize across different domains successfully. However, it is worth noting that there is a more noticeable degradation in performance when the model is pre-trained on the ScanNet dataset and subsequently tested on the Matterport3D dataset. This discrepancy can be attributed to the greater diversity and larger scales of rooms within the Matterport3D dataset compared to the ScanNet dataset. Nonetheless, our model still demonstrates a reasonable level of generalizability across these different datasets and domains.

%% file: src/tables/table_5.tex
\begin{table}[t]
\begin{center}
\resizebox{.9\columnwidth}{!}{
\begin{tabular}{lccc}
\toprule
Supervision type & 
Num. views &
\acc & 
\iou \\
\midrule
2D GT                    & 15     & \textbf{48.2}      & \textbf{32.6}\\
2D GT                    & 5      & 47.9      & 30.1\\
2D GT \textit{w/o VS}    & 0      & 42.1      & 26.9\\
\midrule
2D Pseudo                & 15     & \textbf{37.4}      & \textbf{23.1}\\
2D Pseudo                & 5      & 35.3      & 21.5\\
2D Pseudo \textit{w/o VS}& 0      & 32.1      & 20.5\\
\bottomrule
\end{tabular}
}
\end{center}
\caption{Virtual view selection \textit{(VS)} boosts semantic labeling \textit{(SG)} performance for Matterport3D dataset~\cite{chang2017matterport3d}.}
\label{tbl:ablation-virtual-views}
\end{table}

%% file: src/tables/table_6.tex
\begin{table}[t]
\centering
\resizebox{0.725\columnwidth}{!}{
\begin{tabular}{lcc}
\toprule
Model      & \acc & \iou \\ \midrule
Ours w/o completion & 41.2          & 32.8          \\ 
Ours                & \textbf{48.3} & \textbf{36.7} \\ \bottomrule
\end{tabular}
}

\caption{The effect of geometry completion on semantic segmentation \textit{(SG)} performance.}
\label{tbl:ablation-sg-geometry-completion}
\end{table}

%% file: src/tables/table_7.tex
\begin{table}[t]
\centering
\resizebox{0.75\columnwidth}{!}{
\begin{tabular}{lcc}
\toprule
Method       & \acc & \iou \\ \midrule
Ours w/o color encoder & 40.9          & 22.4          \\
Ours w/o color decoder & 43.5          & 26.3          \\
Ours                   & \textbf{44.8} & \textbf{28.6} \\ \bottomrule
\end{tabular}
}
\caption{The effect of encoding appearance information, via a color encoder and a decoder, on semantic reconstruction \textit{(SSC)} performance.}
\label{tbl:ablation-ssc-color}

\end{table}

%% file: src/tables/table_8.tex
\begin{table}[t]
\centering
\resizebox{0.75\columnwidth}{!}{
\begin{tabular}{lcc}
\toprule
Level of completeness   & \acc     & \iou         \\ \midrule
10--20\%  & 27.5       & 19.1    \\
20--30\%  & 39.9       & 28.6    \\
30--40\%  & 48.3       & 36.7    \\
40--50\%  & 52.1       & 40.4    \\ \bottomrule
\end{tabular}
}
\caption{The effect of varying levels of completeness of the input data on semantic segmentation \textit{(SG)} performance on Matterport3D~\cite{chang2017matterport3d}. }
\label{tbl:sparsity-of-inputs}
\end{table}

%% file: src/main/05-conclusion.tex
\section{Conclusion}
\label{sec:conclusion}

We have presented the first-to-date algorithm to learn geometry completion, scan colorization, and semantic segmentation in a single, self-supervised training procedure. 
Our approach to self-supervised learning builds on several crucial design choices, most importantly, an efficient multi-modal deep neural U-network with residual blocks,
a differentiable rendering technique augmented to produce semantic maps, 
and progress in universal semantic pretraining. 
Fundamentally, our approach enables joint geometry and color reconstruction as well as semantic labeling for unseen scenes without requiring ground-truth labels, a stepping stone in building accurate, semantic models of real-world environments. 
We have additionally established that adding ground-truth information where it is available considerably improves semantic reconstruction performance; in fact, leveraging 3D ground-truth enables achieving state-of-the-art results.

Though in our experiments we have proved that our approach, even with a limited number of RGB-D images captured in a new room, it is capable of generating reconstructed colored meshes with 3D semantic labels. There still exists a performance gap when we apply cross-evaluation on Matterport3D and ScanNet datasets. The substantial disparities in object distributions, room scales, decoration styles, and lighting conditions contribute to this domain adaption issue, although they share some common semantic categories. These challenges commonly arise in real robotic scenarios. Addressing this challenge in future research through the exploration of data augmentation and test-time adaptation techniques holds promise. Furthermore, in the context of large-scale outdoor environments, maintaining a balance between training efficiency and scan resolution remains a challenge that requires further investigation.

\input src/tables/table_9.tex

%% file: src/tables/table_9.tex
\begin{table}[t]
\centering
\resizebox{0.95\columnwidth}{!}{
\begin{tabular}{ccrr}
\toprule
Training Dataset & Testing Dataset  & \acc     & \iou    \\ \midrule
Matterport3D & Matterport3D         & 48.3     & 36.7    \\
Matterport3D & ScanNet              & 36.5     & 27.4    \\
ScanNet & ScanNet                   & 50.8     & 36.4    \\
ScanNet & Matterport3D              & 22.8     & 17.9    \\
\bottomrule
\end{tabular}
}
\caption{Cross-data evaluation of our method on Matterport3D~\cite{chang2017matterport3d} and ScanNet~\cite{dai2017scannet} datasets for the semantic scene completion \textit{(SSC)} task.}
\label{tbl:cross-validation}
\end{table}

%% file: src/main/06-acknowledgement.tex
\section*{Acknowledgements}
\label{sec:ack}
We thank the anonymous reviewers and AC for their
valuable comments. Research presented here has been supported by the ERC Starting Grant Scan2CAD (804724), National Natural Science Foundation of China under Grant Number 62201603, 61921001, the China Postdoctoral Science Foundation under Grant Number 2023TQ0088, the Postdoctoral Fellowship Program of CPSF under Grant Number GZC20233539.
Kai Xu is supported by the NSFC (62325211, 62132021) and the Major Program of Xiangjiang Laboratory (23XJ01009).

%% file: src/biographies.tex
\begin{IEEEbiography}[
{\includegraphics[width=1in,height=1.25in,clip,keepaspectratio]{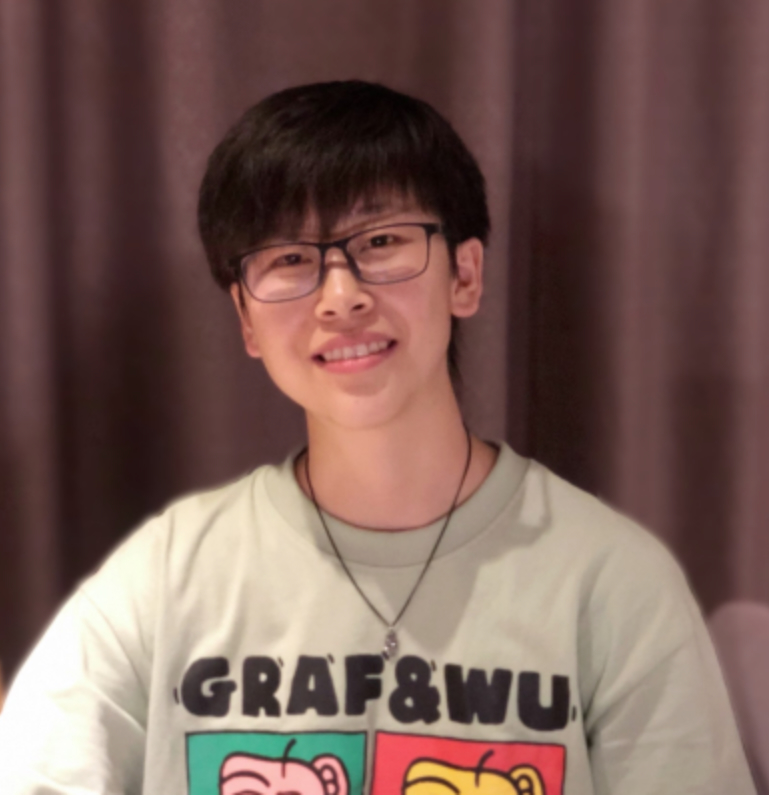}}
]{Junwen Huang}
is a Ph.D candidate in the Chair for Computer Aided Medical Procedures, Technical University of Munich, Germany. She is also a junior member at the Munich Center for Machine Learning. Her research focuses on 3D vision in terms of scene reconstruction and understanding, multi-modal representation learning, point cloud registration, and object 6D pose estimation. 
\end{IEEEbiography}

\begin{IEEEbiography}[
{\includegraphics[width=1in,height=1.25in,clip,keepaspectratio]{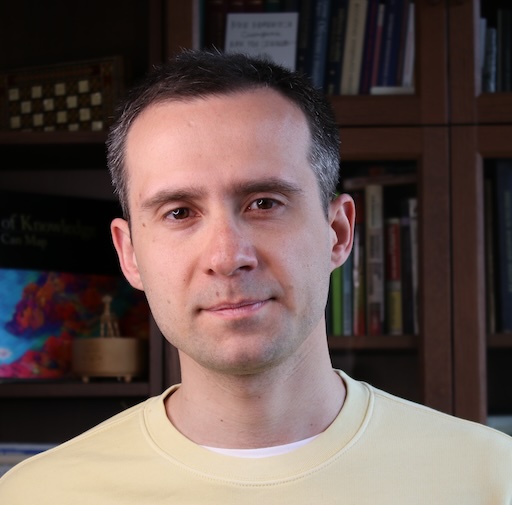}}
]{Alexey Artemov} is a postdoctoral researcher at the Technical University of Munich, focusing on 3D scanning and reconstruction. Alexey obtained his Ph.D. in 2017 from the Institute for Systems Analysis of the Russian Academy of Sciences. In 2021, he received the Ilya Segalovich Award for young researchers.
\end{IEEEbiography}

\begin{IEEEbiography}[
{\includegraphics[width=1in,height=1.25in,clip,keepaspectratio]{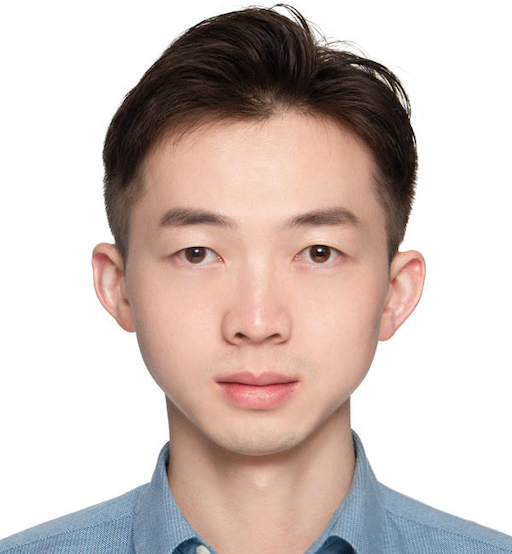}}
]{Yujin Chen}
received the B.Eng. and M.Sc. in Geo-Information from Wuhan University. He is currently a Ph.D. student with the Visual Computing Lab, Technical University of Munich, Germany. His research interests include computer vision and machine learning. His current focus is 3D scene understanding, pose estimation, motion analysis, and representation learning.
\end{IEEEbiography}

\begin{IEEEbiography}[
{\includegraphics[width=1in,height=1.25in,clip,keepaspectratio]{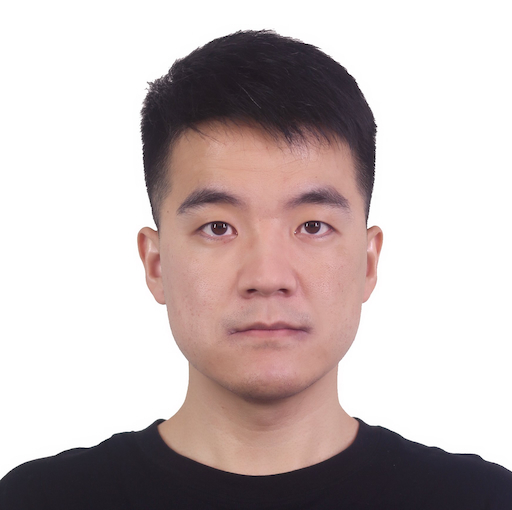}}
]{Shuaifeng Zhi} received his Ph.D. degree in Computing Research at the Dyson Robotics Laboratory, Imperial College London, UK, in 2021. He is currently a Lecturer (Assistant Professor) at the Comprehensive Situational Awareness Laboratory (CSA Lab), College of Electronic Science and Technology, China. He was a 6-month visiting student at 5GIC, University of Surrey, UK, in 2015. 
His current research interests focus on robot vision, particularly on scene understanding, neural scene representation, and semantic SLAM.
\end{IEEEbiography}

\begin{IEEEbiography}[
{\includegraphics[width=1in,height=1.25in,clip,keepaspectratio]{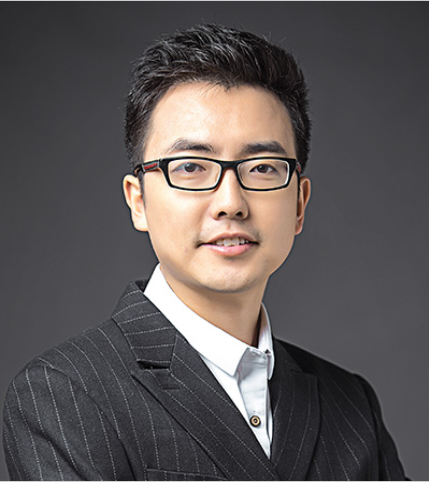}}
]{Kai Xu}
is a professor at the Xiangjiang Laboratory, China. He conducted visiting research at Simon Fraser University and Princeton University. His research interests include geometric modeling and shape analysis, especially on data-driven approaches to the problems in those directions, as well as 3D vision and its robotic applications. He has published over 80 research papers, including 20+ SIGGRAPH/TOG papers. He has co-organized several SIGGRAPH Asia courses and Eurographics STAR tutorials. He serves on the editorial board of ACM Transactions on Graphics, Computer Graphics Forum, Computers \& Graphics, and The Visual Computer. He also served as program co-chair of CAD/Graphics 2017, ICVRV 2017 and ISVC 2018, as well as PC member for several prestigious conferences including SIGGRAPH, SIGGRAPH Asia, Eurographics, SGP, PG, etc. His research work can be found in his personal website: www.kevinkaixu.net
\end{IEEEbiography}

\begin{IEEEbiography}[
{\includegraphics[width=1in,height=1.25in,clip,keepaspectratio]{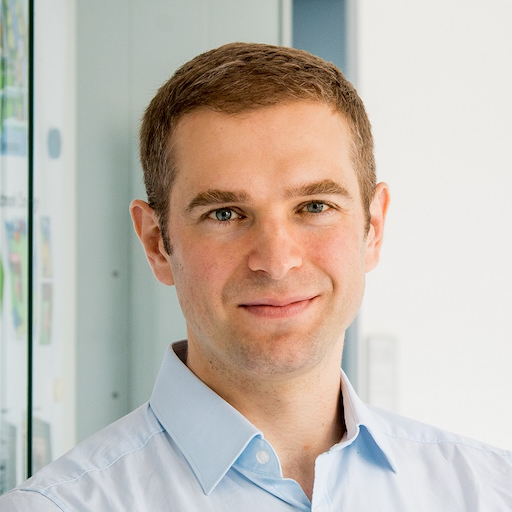}}
]{Matthias Nie{\ss}ner} is a Professor at the Technical University of Munich, where he leads the Visual Computing Lab. Before, he was a Visiting Assistant Professor at Stanford University. Prof. Nießner’s research lies at the intersection of computer vision, graphics, and machine learning, where he is particularly interested in cutting-edge techniques for 3D reconstruction, semantic 3D scene understanding, video editing, and AI-driven video synthesis. In total, he has published over 150 academic publications, including 25 papers at the prestigious ACM Transactions on Graphics (SIGGRAPH / SIGGRAPH Asia) journal and 55 works at the leading vision conferences (CVPR, ECCV, ICCV); several of these works won best paper awards, including at SIGCHI’14, HPG’15, SPG’18, and the SIGGRAPH’16 Emerging Technologies Award for the best Live Demo.
\end{IEEEbiography}


